\documentclass{article}

\PassOptionsToPackage{numbers, sort&compress}{natbib}

\usepackage[dblblindworkshop, final]{neurips_2026}

\usepackage{enumitem}
\usepackage[utf8]{inputenc} 
\usepackage[T1]{fontenc}    
\usepackage{hyperref}       
\usepackage{wrapfig}
\usepackage{url}            
\usepackage{booktabs}       
\usepackage{amsfonts}       
\usepackage{amsmath}
\usepackage{cleveref}
\crefname{figure}{Figure}{Figures}   
\Crefname{figure}{Figure}{Figures}
\usepackage{nicefrac}       
\usepackage{microtype}      
\usepackage{xcolor}         
\usepackage{graphicx}       

\title{From States to Dynamics: Detecting Rule Violations in OthelloGPT}
\workshoptitle{Interpretability as a Science}

\author{%
  David L.~Cao \\
  Harvard College\\
  Harvard University\\
  Cambridge, MA 02138 \\
  \texttt{dcao@college.harvard.edu} \\
  \And
  Sai Sumedh R.~Hindupur \\
  School of Engineering and Applied Sciences\\
  Harvard University\\
  Cambridge, MA 02138 \\
  \texttt{sumedh.hindupur@gmail.com} \\
  \And
  Demba Ba \\
  School of Engineering and Applied Sciences\\
  Kempner Institute, Harvard University\\
  Cambridge, MA 02138 \\
  \texttt{deb884@g.harvard.edu} \\
}

\begin{document}

\maketitle

\begin{abstract}

Sequence models trained on game moves have been shown to encode world models, with structured linear probes revealing the underlying game states. World models, however, are inherently dynamic: in deterministic board games, a set of rules constrain how states evolve over time. We therefore ask if and how OthelloGPT, a transformer trained only on Othello move sequences, manifests the rules of the game in its activation space. We divide the rules of Othello, and symmetrically their violations, into three classes: placing a disc on an empty square (R1), flanking an opponent's disc (R2), and moving in turn (R3). We find that linear probes accurately separate legal from illegal moves, and also distinguish between the three kinds of rule violations. We introduce an unsupervised rule-violation detector based on the step length of activations between consecutive moves, which identifies illegal moves above chance. We further improve our detection metric by computing step length in the latent space of select sparse autoencoders. Together, we find that OthelloGPT follows game rules of Othello to a large extent, and that we can detect rule violations using structured detectors on model activations.\footnote{Code is available at \url{https://github.com/dcowdcao/states-to-dynamics}.}

\end{abstract}

\vspace{-1em}
\section{Introduction}


How well does a deep neural network understand structure and dynamics of its data? In an effort to address this question, much of the current work has studied how toy game-playing models encode \emph{state-based} concepts~\citep{li2023emergent, nanda2023emergent, toshniwal2022chess, karvonen2024chess, ivanitskiy2023linearly, lovering2022hex}. \citet{li2023emergent}, for instance, trained OthelloGPT, which predicts the next move from a sequence of Othello moves, to study how a model encodes a world model of its training data. Follow-up studies have since shown that, by using linear probes~\citep{nanda2023emergent, hazineh2023linear, karvonen2024chess, yuan2025revisiting, lee2026tpr, chawla2026metaothello} and sparse autoencoders (SAEs)~\citep{huben2024research, he2024dictionary, karvonen2024measuring}, we can read out the board state of OthelloGPT and of similar chess-playing models, and that we can causally edit the model's next move by altering its internal representation of the board~\citep{li2023emergent, nanda2023emergent, hazineh2023linear, davis2026probes}.

However, decoding state-based concepts alone tells us little about how models actually capture the structure of their data. \citet{vafa2024evaluating} show that, unlike OthelloGPT trained on synthetic games, OthelloGPT trained on real championship games does not treat the same board as the same state (it predicts a next move after one opening that it rejects after another) and produces mostly invalid games when even a small fraction of its moves is replaced with other legal moves. This tells us that a model's next-move legality and board-state decoding can both be near-perfect while the model is likely memorizing in-distribution sequence statistics. Therefore, to understand how game-playing models truly capture structure in their data, we need to understand not only how they represent board states but also the \emph{dynamics} of the game and when their generations violate underlying rules. 

Several studies have examined the dynamics of representations within models. For instance, \citet{lubana2026priors} introduce temporal SAEs, which decompose activations into a component predictable from prior context and a residual carrying novel information, thereby separating slow-moving from fast-moving concepts. Building on the perceptual straightening hypothesis~\citep{henaff2019straightening}, \citet{interno2025restrav} develop a representation space based on the temporal curvature and stepwise distance of a vision transformer's frame-by-frame representations, and find that simple classifiers on it distinguish AI-generated from real videos. Transformers trained on hidden Markov processes like Mess3~\citep{marzen2017nearly} linearly represent the Bayesian belief state in their residual stream~\citep{shai2024belief}, summing small corrections from each prior token in the attention blocks to form a fractal-like point in the probability simplex~\citep{piotrowski2025constrained}. Finally, \citet{edelman2024induction, park2025competition} show that transformers trained on a finite set of Markov chains can infer the transition structure of unseen chains in context.

Within game-playing models, specifically OthelloGPT, current studies have looked at rule dynamics. For instance, \citet{lin2024heuristics} find a circuit for the flanking rule in which a neuron in MLP~1 fires after a flanking move and updates the model's board representation to flip the opponent's discs, and that, under the logit lens, legal moves, treated as a single class, are gradually promoted across layers. Additionally, \citet{singh2025rule} fit a regression decision tree to each MLP neuron from ground-truth board features, giving each neuron a disjunctive normal form (DNF) description; given a rule query, their tool automatically surfaces the neurons whose DNF evaluates to true, and they find that rule-based neurons implementing the flanking condition are concentrated in layers 5--6.

This line of work has primarily focused on the flanking-and-flipping rule and identifying individual neurons and circuits that implement this specific state update. However, when game-playing models generate longer trajectories of moves, to what extent do they respect the rules of the underlying game? When there are rule violations, can we detect these violations from the model's internals? 
  
Specifically, in this work, we address three questions around rule dynamics. \textbf{First}, how well does OthelloGPT follow the rules of Othello? \textbf{Second}, when OthelloGPT makes an illegal move, can we detect the illegal move from the model's activations? \textbf{Third}, since SAEs have been shown to recover board states~\citep{huben2024research, karvonen2024measuring, davis2026probes}, can SAE latents help us detect rule violations better than model activations. We make three main contributions:

\begin{itemize}[leftmargin=16pt, itemsep=1.5pt, topsep=1pt, parsep=1.5pt, partopsep=1pt]

    \item \textbf{Supervised legal/illegal move identification.} We train linear probes on OthelloGPT's residual stream activations to classify legal and illegal moves, as well as the types of illegal moves. We show that these probes are capable of accurately separating legal from illegal moves and each type of rule violation.
    \item \textbf{Unsupervised legal/illegal move identification.} We develop a step-length metric that detects whether a move is illegal. We show that this metric detects, in an unsupervised manner, the moves at which a violation occurred above chance.
    \item \textbf{Certain SAEs improve legal/illegal move identification.} We find that applying our detector to the latent activations of select SAEs improves our legal/illegal move detector. 
\end{itemize}

\vspace{-1em}
\section{Preliminaries}

\textbf{Othello and OthelloGPT.} Othello is a two-player game played on an $8 \times 8$ board with black and white discs. The game starts from four discs on the four center squares, and black moves first; the game ends when neither player can move, and the player with the most discs of their color on the board wins. The maximum number of moves in a game is $60$. 

In this work, we analyze OthelloGPT~\citep{li2023emergent}, an 8-layer decoder-only transformer ($d_{\text{model}} = 512$, 8 heads, 25.3M parameters) trained on 20M synthetic games of uniform-random legal play to
predict the next move from a sequence of move tokens alone. Throughout this study, we read its residual stream after layer 5, because it has high accuracy (see
Appendix A.1).

\textbf{Linear probes \& Sparse Autoencoders.} A linear probe is a \emph{supervised} method that identifies directions in model activations for classes of interest~\citep{alain2016probes}. 
SAEs are an unsupervised representation learning method~\citep{bricken2023monosemanticity, cunningham2023sparse} to represent activations as a sparse linear combination of an over-complete basis of interpretable dictionary atoms~\citep{olshausen1997sparse, elhage2022toy}.  The goal of SAEs is monosemanticity: a one-to-one mapping between learnt features and interpretable concepts.

\vspace{-1em}
\section{OthelloGPT and the Rules of Othello}

In this section, we introduce the rules of Othello, define the $3$ classes of illegal moves, and show that OthelloGPT follows these rules.


\begin{wrapfigure}[20]{tR}{0.6\textwidth}
  \centering
  \includegraphics[width=0.57\textwidth]{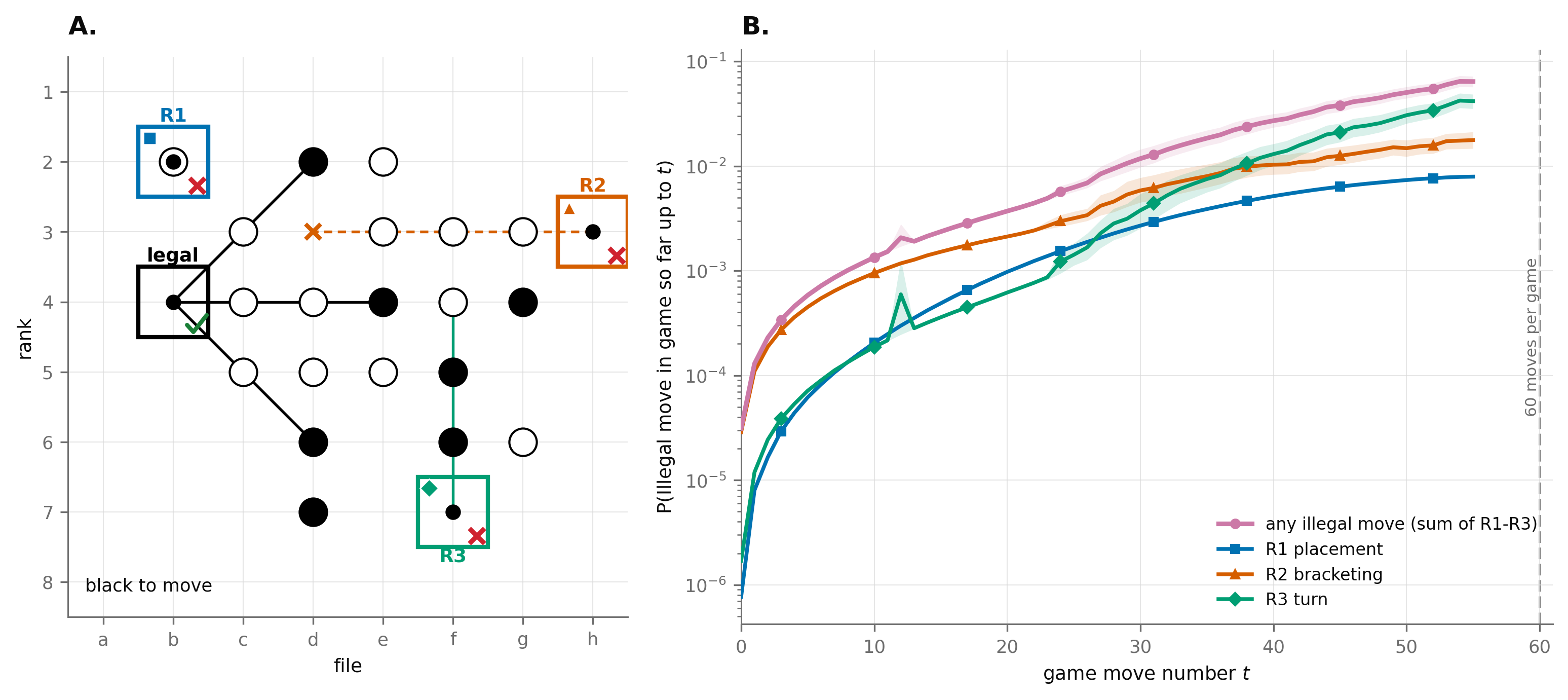}
    \caption{\textbf{OthelloGPT largely follows the rules of Othello.}
    \textbf{(a)} Board state from a randomly chosen game at move $t=16$, with black to move, illustrating a legal move and the three classes of rule violations (R1, R2, R3).
    \textbf{(b)} Next-token probability of generating an illegal move at move $t$, separated by violation class and averaged over $n=1000$ games. This probability remains low throughout the game, reaching $7.1\%$ at $t=58$, indicating that OthelloGPT generally generates moves consistent with the rules of Othello.}
  \label{fig:move_generation}
\end{wrapfigure}

The game rules of Othello, and the three illegal move classes they simultaneously define, are summarized as follows:
\begin{itemize}[leftmargin=16pt, itemsep=1.5pt, topsep=1pt, parsep=1.5pt, partopsep=1pt]
    \item \textbf{R1 (Empty-Only):} The player must place a disc on a currently empty square. Illegal moves in class R1 \textit{place discs on squares already occupied.}
    \item \textbf{R2 (Flanking):} The player's disc must flank a contiguous line of their opponent's pieces (horizontally, vertically, and/or diagonally). After the move, the opponent's discs turn to the player's color. Illegal moves in class R2 are placed on empty squares but \emph{do not flank the opponent's discs.}
    \item \textbf{R3 (Turn):} The player's move is only valid if it is the player's turn. Players take turns moving unless a player cannot play a move that flanks their opponent's discs (per R2), in which case their turn is skipped. Illegal moves in class R3 \textit{would have been legal} if it were actually the opponent's turn. Note that R3 is a subclass of R2. 
    
\end{itemize}

See \cref{fig:move_generation}(a) for examples of R1, R2, and R3 in play. In \cref{fig:move_generation}(b), we find that the next-token probability of OthelloGPT playing an illegal move remains low throughout the game (under $\sim 7\%$ over the $58$ moves that OthelloGPT can predict, since the model has context $59$ tokens), although it increases with move number.

As demonstrated in prior work ~\citep{hazineh2023linear, li2023emergent, nanda2023emergent, lee2026tpr}, board state can be decoded from the model's activations, which we demonstrate for tensor product probes (TPRs) and SAEs in App. \cref{fig:sae_sweep}. \cref{fig:flanking}.

\section{Linear Probes Distinguish Legal \& Illegal Moves} \label{sec:4}


\begin{figure}[htbp]
  \centering
  \includegraphics[width=0.7\linewidth]{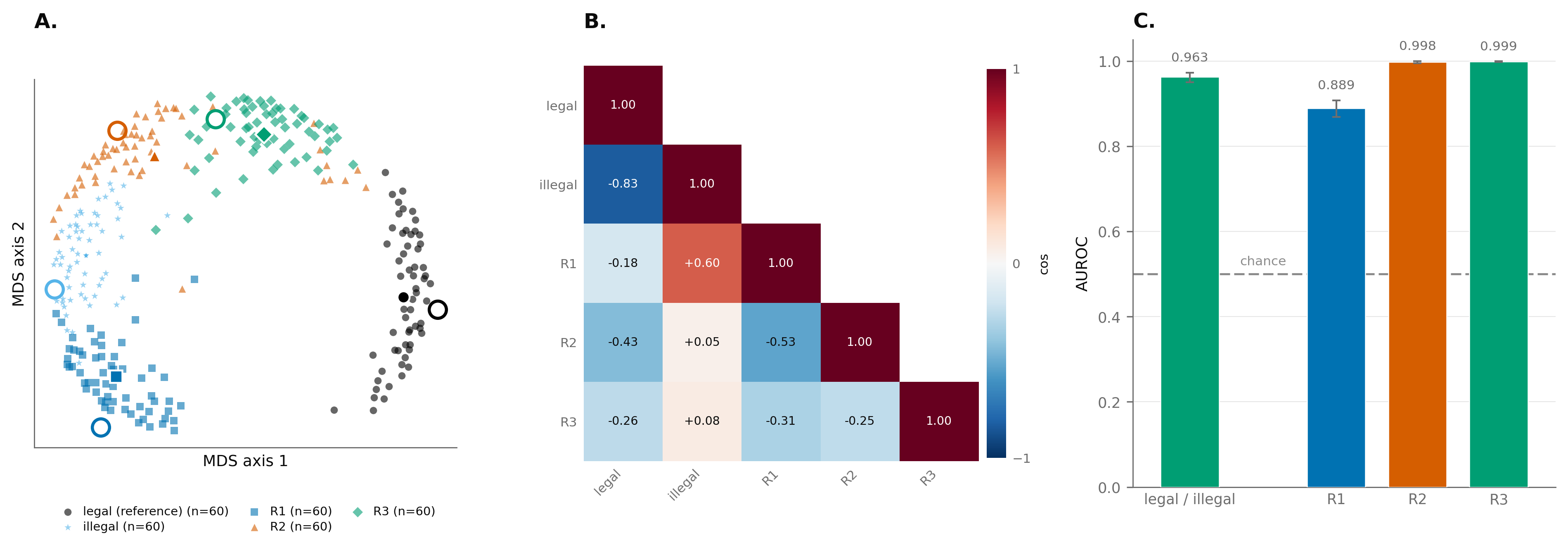}
  \caption{\textbf{Rule violations can be distinguished using linear probes.} \textbf{(a)} 2D Multidimensional Scaling (MDS) visualization of linear probe directions (ring is linear probe direction, point is re-fitted probe per square) of the five classes (legal, illegal, R1, R2, and R3)  \textbf{(b)} Cosine similarity matrix between the five probe directions. Illegal moves are anti-correlated with legal moves, and illegal classes are also anti-correlated with each other. \textbf{(c)} AUROC of linear probes for distinguishing each class. Performance is highest for distinguishing legal from illegal moves and lowest for distinguishing legal from R1 violations. }
  \label{fig:MDS_cosine_probe}
\end{figure}


In this section, we first show that the rule violations of Othello can be accurately decoded from the model's residual stream activations using two linear probes: the first is trained to classify moves as $\{\rm legal, \rm illegal\}$, and the second, which further classifies the rule violations in illegal moves -- see \cref{fig:MDS_cosine_probe} (c).

Second, we visualize the five aggregate linear probe directions under specific rule violations via 2D Multidimensional Scaling (MDS)~\citep{kruskal1964mds}; see \cref{app:probes} for a detailed description of the MDS construction. As is evident in \cref{fig:MDS_cosine_probe} (a), probes trained on OthelloGPT accurately distinguish between legal and the three classes of rule violations.

Third, cosine similarity between the probe directions demonstrates how similar or dissimilar each of the five directions are to each other. For example, in \cref{fig:MDS_cosine_probe} (b), $\{\rm legal, \rm illegal \}$ scores $-0.83$, indicating that the linear directions used to distinguish these classes are nearly opposite in activation space. Additionally, we find that \{R1, R2, R3\} all score $< 0$ with each other. This makes sense because, for instance, if an illegal move falls under violation R1, it cannot simultaneously fall under R2 or R3. 

\vspace{-1em}
\section{Unsupervised Illegal Move Detection}


\begin{figure}[t]
  \centering
  \includegraphics[width=0.8\linewidth]{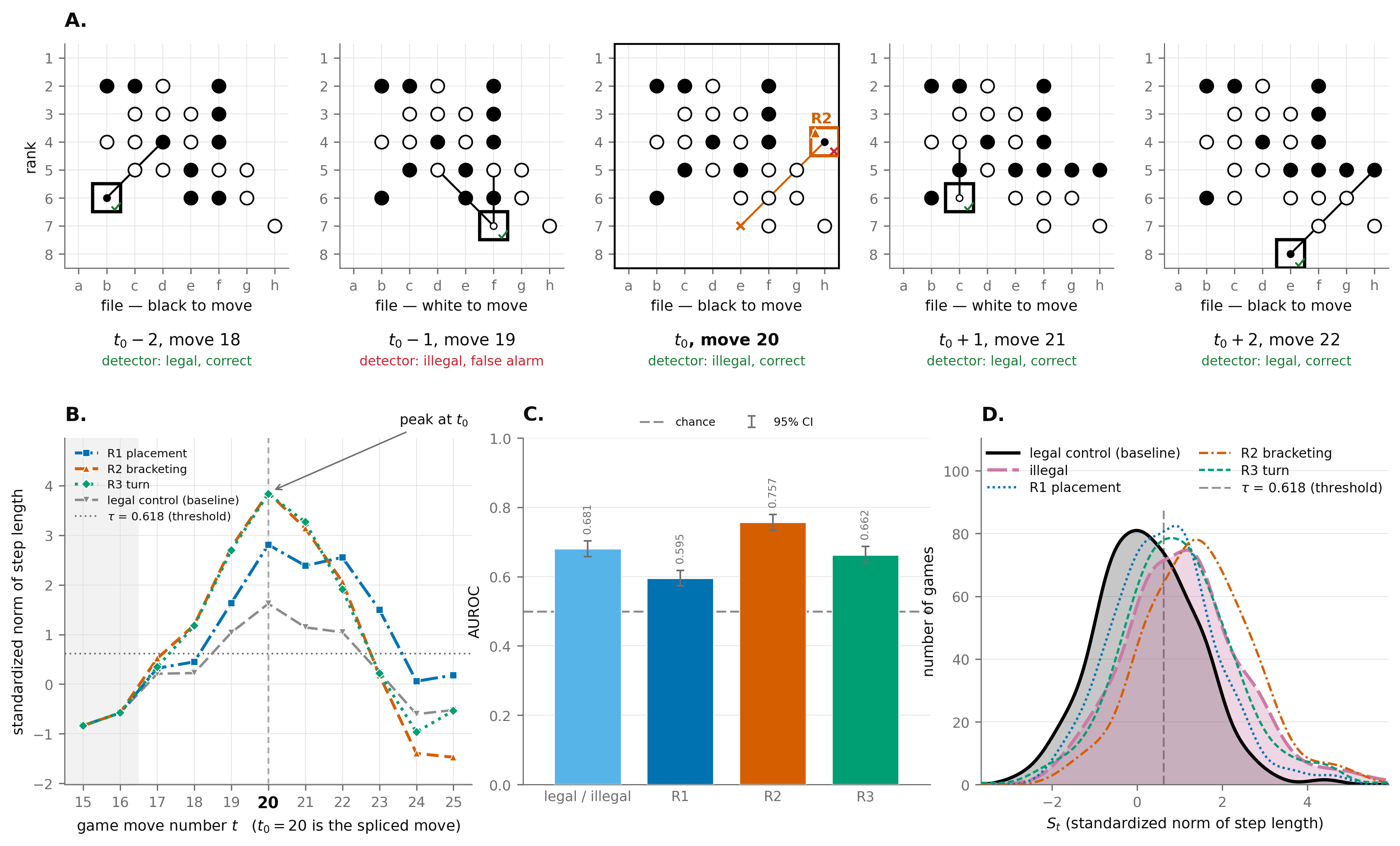}
    \caption{\textbf{Step-length metric detects rule violations in residual stream activations.}
    \textbf{(a)} Example Othello game centered at $t_0 = 20$, where an R2 violation occurs; the metric correctly classifies $4/5$ moves in the displayed window.
    \textbf{(b)} The detection metric computed from residual stream activations peaks at the illegal move $t_0$.
    \textbf{(c)} AUROC for detecting $\{\rm illegal, R1, R2, R3\}$ against $\{\rm legal\}$ exceeds the random baseline of $0.50$ for all violation classes.
    \textbf{(d)} Despite this separation, the metric distributions show substantial overlap between legal moves and illegal moves, including each rule-violation class.}
  \label{fig:detector_example}
  \vspace{-1em}
\end{figure}


In this section, taking inspiration from ~\citep{interno2025restrav}, we propose a \emph{unsupervised} metric that detects whether a move in a game violated a rule. We first apply the metric to the model's residual stream activations, and then show that the metric is improved under select SAE latent activations. 

\vspace{-0.5em}
\subsection{Detection Using Residual Stream Activations}

Let $z_t \in \mathbb{R}^{512}$ denote OthelloGPT's layer-$5$ (see App. \cref{fig:probe_layer}) residual stream activations at the position of move $t$ and let $d_t$ be the norm step length at move $t$:

\begin{equation}
  d_t = \lVert z_t - z_{t-1} \rVert_2.
\label{eq:step}
\end{equation}

We standardize $d_t$ for each $t \in [2, 58]$ against a corpus of $n = 2000$ legal games. Thus, we have $g_t = -\,\frac{d_t - \mu_t}{\sigma_t}$ (where $\mu_t, \sigma_t$ are the mean and standard deviation over multiple games of $d_t$, respectively), and we use the moving average of $g_t$ over a window of $m$ moves before and after $t$, i.e. $[t-m-1, t+m]$, for $m = 4$: $S_t \;=\; \sqrt{\tfrac{m}{2}}\left(\frac{1}{m}\sum_{i=0}^{m-1} g_{t+i}\;-\;\frac{1}{m}\sum_{i=1}^{m} g_{t-i}\right)$.

We use a moving sum because we are asking ``was there a rule violation at $t$?'', and we answer this question by measuring the effect that $t$ has on $g_t$ in a window of $m = 4$ moves before and after $t$.

In \cref{fig:detector_example}, we show our metric, $S_t$, in action on a specific example game. In that game, we find that $S_t$ peaks at $t_0$, the illegal move number, and correctly classifies $4/5$ of the window of moves as $\rm illegal$ or $\rm legal$ -- see (a) and (b). 

We also measure the AUROC of our metric $S_t$ for detecting $\{\rm illegal,\rm R1, \rm R2, \rm R3\}$ against $\{\rm legal\}$. See \cref{fig:detector_example} (c) -- we find that our metric achieves an AUROC of $\sim 0.68$ for detecting illegal moves, which is considerably better than the random baseline of $0.50$. Since AUROC is the probability that a randomly picked violation scores higher than a randomly picked legal move, we do not need to choose threshold(s) for $S_t$. Among specific rule violations, our metric performs best at detecting $\rm R2$ violations, achieving an AUROC of $\sim 0.76$, and performs above chance at detecting $\rm R1$ violations, achieving an AUROC of $\sim 0.60$. 

\vspace{-0.5em}
\subsection{Improved Detection Using SAEs}

\begin{figure}[t]
  \centering
  \includegraphics[width=0.8\linewidth]{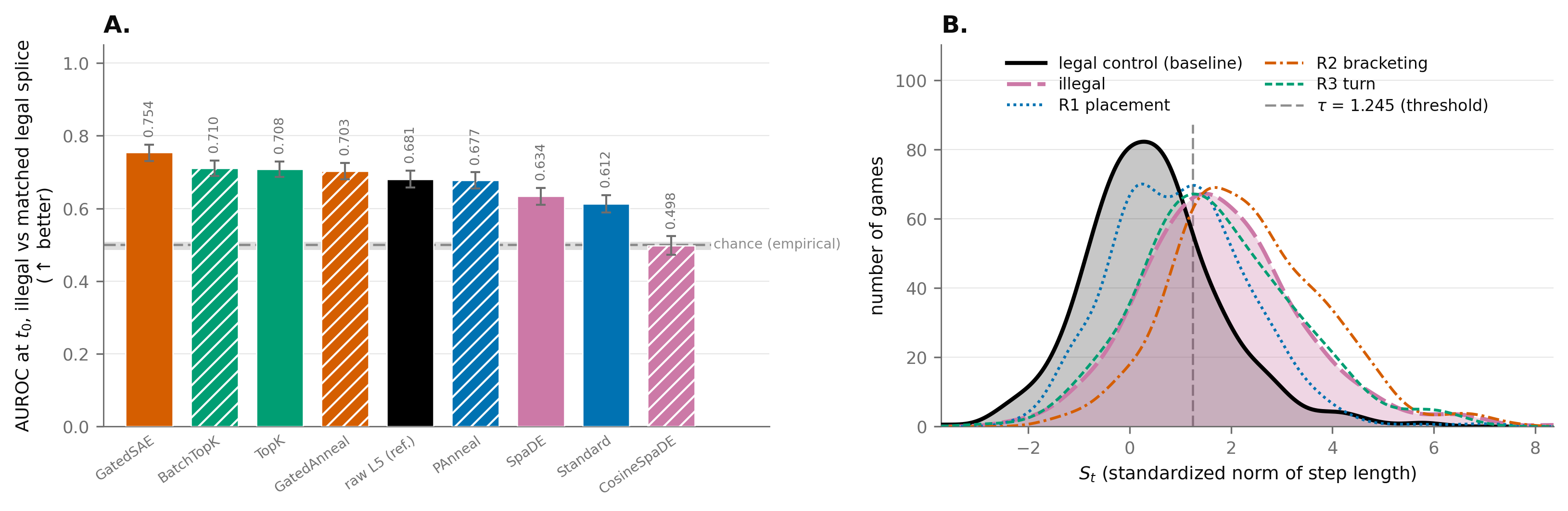}
  \caption{\textbf{Detecting rule violations in SAE latent space.} \textbf{(a)} AUROC of the detector applied in the latent spaces of eight SAE architectures. Several SAE representations outperform the raw residual stream, with GatedSAE achieving the highest AUROC ($\sim 0.75$) for distinguishing illegal from legal moves. \textbf{(b)} Less overlap in the distribution of the metric between legal moves and illegal moves is observed for GatedSAE latent space as opposed to that of residual stream activations (see \cref{fig:detector_example}).}
  \label{fig:detector_SAE}
  \vspace{-1em}
\end{figure}

While our detector performs reasonably well in the residual stream, we next ask whether SAE representations provide a space in which rule violations are more easily separated. We evaluate eight SAE architectures: Standard SAE~\citep{bricken2023monosemanticity}, p-Anneal~\citep{karvonen2024measuring}, GatedSAE~\citep{rajamanoharan2024gated}, Gated+p-Anneal~\citep{karvonen2024measuring}, TopK~\citep{makhzani2013ksparse, gao2025scaling}, BatchTopK~\citep{bussmann2024batchtopk}, SpaDE~\citep{hindupur2025projecting}, and Cosine SpaDE, a cosine-similarity variant of SpaDE. See App. \cref{app:SAE_descriptions}. Among them, GatedSAE performs best, achieving an AUROC of $\sim 0.75$ for distinguishing illegal from legal moves (\cref{fig:detector_SAE}a).


\vspace{-0.5em}
\section{Discussion \& Limitations}

In this work, we investigate whether and how OthelloGPT follows the rules governing the dynamics of Othello from move sequences alone. We first show that OthelloGPT largely follows the rules during generation with high accuracy throughout the game. Second, we find that linear probes, a supervised method, distinguish between legal from illegal moves and the three classes of rule violations, showing that move legality is linearly decodable from model activations. Finally, we show that these rule violations can be detected with an unsupervised metric based on the step-length of model activations, and that our detector is improved under some SAE latent activations. Together, these results suggest that OthelloGPT largely follows the rules of Othello and rule violations can be detected from the model's activations.

\paragraph{Limitations} First, while our detector detects illegal moves, there remains room for improvement in the detector design, as we see significant overlap between the step length of legal and illegal moves (see \cref{fig:detector_example} (d) and \cref{fig:detector_SAE} (b)). Second, we show that step length is correlated with illegal moves. We defer causal claims to future work: can we perturb the model's activations towards an illegal move for each type of rule violation? Finally, our detector can only classify legal and illegal moves. We see an avenue for future work to build upon our metric to develop a metric that also classifies the specific rule violations. 


\begin{ack}

This work has been made possible in part by a gift from the Chan Zuckerberg Initiative Foundation to establish the Kempner Institute for the Study of Natural and Artificial Intelligence at Harvard University. Additionally, DC thanks the Kempner Institute for access to the cluster and compute resources, and all authors thank the CRISP Group at Harvard SEAS for insightful conversations. 
\end{ack}


\newpage
{
\small
\bibliographystyle{plainnat}
\bibliography{refs}   
}


\newpage
\appendix

\section{Experimental Setup}

Every experiment runs on NVIDIA H100 80GB GPUs. Training the sparse autoencoders takes 8--16
hours; every probe fit, rollout and read-out below takes minutes on one GPU. The code is PyTorch
2.5.1 with TransformerLens 1.17.0 under Python 3.11.

\subsection{Model and activations}

We study OthelloGPT~\citep{li2023emergent}, an 8-layer decoder-only transformer trained to
predict the next legal move from a sequence of move tokens alone -- it never sees a board. We
use the TransformerLens~\citep{nanda2022transformerlens} port \texttt{Baidicoot/Othello-GPT-Transformer-Lens}: $d_{\mathrm{model}} =
512$, 8 heads of dimension 64, MLP width 2048, GELU, context length 59, vocabulary 61, 25.3M
parameters. It is loaded with TransformerLens defaults, so LayerNorm is folded into the
adjacent weights and the writing and unembedding weights are centered; every activation below
comes from that folded model. Token 0 is the pass token and tokens 1--60 are the playable
squares, which are the board minus the four center squares -- those start occupied and are
never played.

\begin{figure}[htbp]
  \centering
  \includegraphics[width=0.7\linewidth]{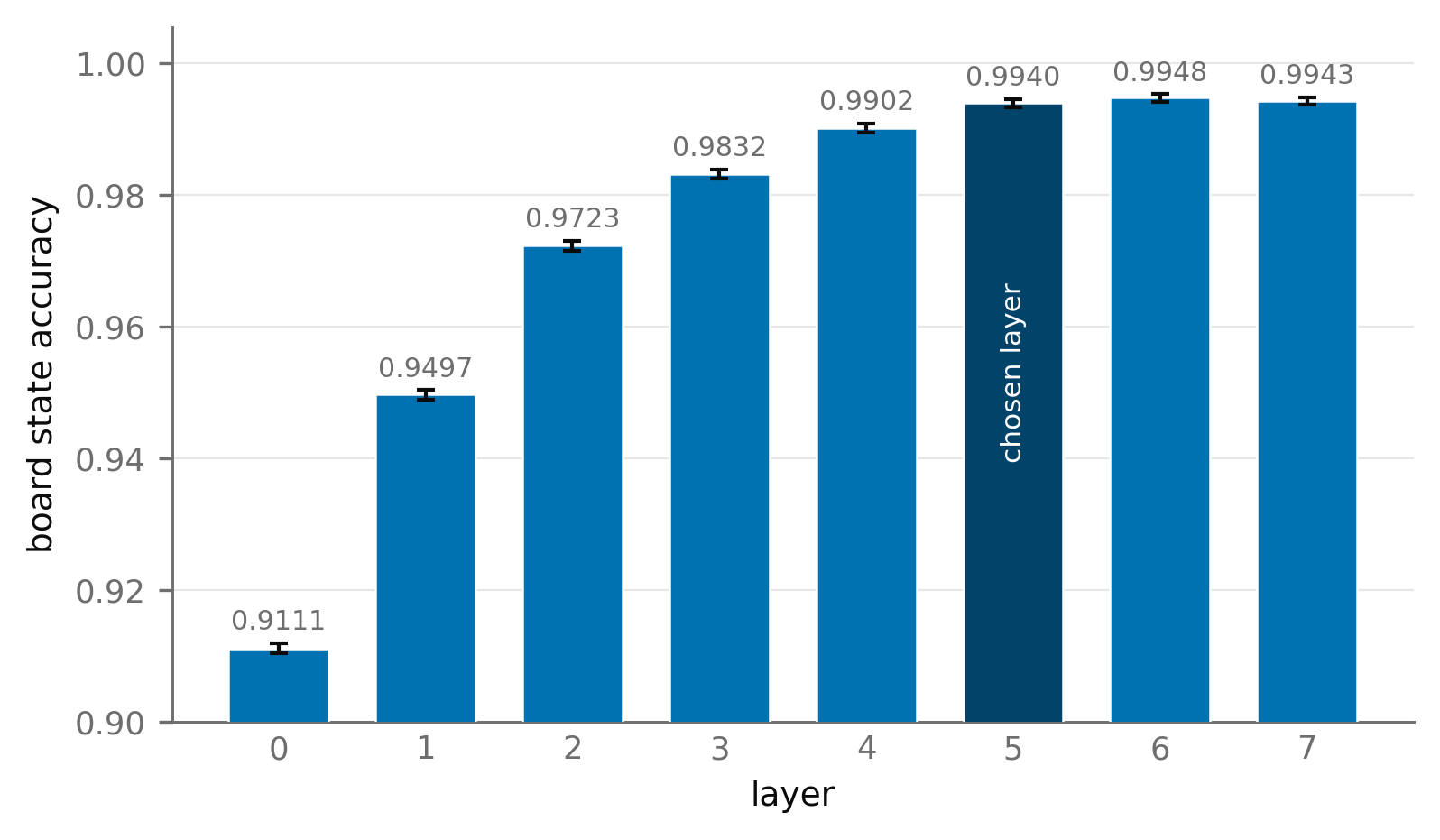}
  \caption{\textbf{Accuracy of linear probe trained on Othello board state over each residual stream layer of OthelloGPT.} A linear probe classifies each of the $64$ squares on the board as $\{\rm empty, \rm mine, \rm theirs\}$ from the \textit{residual stream activations} after each block $\ell \in \{0, \ldots, 7\}$, scored on $n = 1000$ held-out games ($58{,}000$ positions). }
  \label{fig:probe_layer}
\end{figure}

Unless stated otherwise, activations are the residual stream after block 5,
\texttt{blocks.5.hook\_resid\_post}, so $z_t \in \mathbb{R}^{512}$. We read at layer 5 because the board probe is
flat across the top of the stack (\cref{fig:probe_layer}) -- 0.9940 at layer 5 against 0.9948 at layer 6
and 0.9943 at layer 7, a spread of 0.0008 -- so the choice costs nothing and every study shares
it. Position 0 is an attention sink whose norm is $\approx 300$ against the $\approx 23$ of
every other position, so every trajectory measurement drops it: a game contributes 58
positions and 57 steps.


\subsection{Game corpus}

All probes, rollouts and read-outs stream \texttt{adamkarvonen/othello\_45MB\_games}~\citep{karvonen2024measuring}: 990,623
games (792,498 train, 198,125 test), each exactly 60 plies with no pass token. The corpus is
synthetic uniform-random legal play. We group games by $k$-ply prefix and test each prefix's
children against uniform over the engine-legal set. At $k = 5$, no prefix with $n \ge 400$
rejects at $p < 10^{-3}$.

Two properties of the corpus reach every number below. First, games are drawn with
\texttt{shuffle(seed=42, buffer\_size=10{,}000)} over a streaming split that is a single shard,
so the shuffle permutes only a 10,000-row reservoir and drawing $N$ games reads the first
$N + 10{,}000$ rows. It is a window on the corpus, not a sample of it. Every study therefore walks the same leading percent or two of the
corpus in the same order, and their game sets are nested prefixes rather than independent
draws; a different seed replaces $\approx 80\%$ of the games but stays inside that same slice. A
different seed is not a replication.
Where two studies must not share games, we exclude by exact token match rather than by
reseeding, and say so below. Second, the corpus is lexicographically sorted and truncated to
its 45\,MB byte budget, which cuts one opening: \texttt{e6} holds 10.78\% of games where the
board's symmetry group requires 25\% ($\chi^2 = 106{,}901$ on 3 df over the pooled corpus). Nothing in the corpus is
illegal and play off that branch is uniform random-legal, but a per-square aggregate can
inherit an asymmetry that has nothing to do with the model.

Forfeits are unmarked in the move stream -- the same color moves twice -- so whose turn it
is always comes from replay through Li et al.'s board implementation~\citep{li2023emergent}, never from the
parity of the move index.

The sparse autoencoders are trained on the separate stream that Karvonen's pipeline uses~\citep{karvonen2024measuring},
\texttt{taufeeque/othellogpt}.

\subsection{The illegal-move taxonomy}

Given a board and the player to move, each of the 60 vocabulary squares falls into one of seven
categories: \texttt{legal}, and six violations ordered by how much board computation is needed
to detect them. \textbf{R1 placement} is $\{$\texttt{occupied\_own}, \texttt{occupied\_opp}$\}$;
\textbf{R2 bracketing} is $\{$\texttt{empty\_detached}, \texttt{empty\_own\_only},
\texttt{empty\_no\_bracket}$\}$; \textbf{R3 turn} is $\{$\texttt{opponent\_only}$\}$, a square
that flips only for the player who is not to move. The classifier agrees with the engine's legal
set exactly on non-forfeit positions, and we assert that at every position we accept. Forfeit
positions are skipped: there the engine falls back to letting opponent-flipping squares be
played. R2 has three members and R3 has one, so any average over a rule states its $n$.

\subsection{Probes} \label{app:probes}

Two families of linear read-out sit on $z_t$ at layer 5: board-state probes, which report what is
on each square, and legality probes, which report whether the move at that position is legal and
which rule it breaks. Both take labels from engine replay, are fit on the seed-42 stream, and
split \emph{over games}, so no game straddles a split.

\textbf{Board state.} We use a tensor product probe (TPR)~\citep{lee2026tpr},
$\mathrm{logit}(s, c) = r_s^\top M(h) f_c$, at role dimension 52 and filler dimension 2 over the
64 squares and the three mover-relative colors $\{$empty, theirs, mine$\}$: a $[512, 52, 2]$
binding map, a $[64, 52]$ role table and a $[3, 2]$ filler table, 56,582 parameters in total,
57.6\% of the 98,304 of the per-square linear read-out it is measured against. Architecture and loss follow~\citep{lee2026tpr}; the recipe is ours, and both probes run behind one protocol so
that a head-to-head compares probe class and nothing else. Both are fit on 20,000 corpus games at
58 positions each, 1,160,000 rows, split $70/15/15$ into 812,000 training and 174,000 validation
and test rows. We use AdamW at learning rate $10^{-2}$, batch 8192, at most 400 epochs, stopping
early on validation accuracy with patience 20 and minimum improvement $10^{-4}$, and keep the
best-validation weights. Weight decay is $10^{-4}$ for the linear arm; for the TPR we re-select
it on validation over $\{10^{-4}, 10^{-3}, 10^{-2}, 10^{-1}, 0.5, 1.0, 2.0\}$, which picks
$10^{-3}$ (validation 0.99202, against 0.99107 at 2.0). At layer 5 the held-out test accuracies
are 0.9920 (TPR, best epoch 31) and 0.9937 (linear).

\textbf{Reading the TPR.} From the three logits at a square $s$ we plot $\ell_{\mathrm{mine}} -
\ell_{\mathrm{theirs}}$, which is positive when the square is owned by the player to move, against
$\tfrac{1}{2}(\ell_{\mathrm{mine}} + \ell_{\mathrm{theirs}}) - \ell_{\mathrm{empty}}$, which is positive when
it is occupied. We read logit differences rather than filler coordinates because the bilinear
form is gauge-invariant under $f \mapsto Af,\; M \mapsto MA^{-1}$: the measured filler angle
moves from $131.3^\circ$ to $104.5^\circ$ across seeds at constant accuracy.

\textbf{Legality.} We fit two linear read-outs of $z_t$ at the position of the candidate move: a binary
$\{$legal, illegal$\}$ head and a four-way $\{$legal, R1, R2, R3$\}$ head. We mine 800 games at
seed 42. An item is a (game, move index) pair carrying the move actually played plus one
substituted illegal move per constraint rule that has a candidate there; all rows of an item
share \texttt{tokens[:$t$]} and differ in exactly one token, so class is the only thing that
varies. The illegal move is drawn uniformly over that rule's candidate tokens, and positions run
over $1 \ldots 58$. The split is $60/20/20$ over games (480/160/160), giving 27,743 rows per
class for the binary head and 23,641 for the four-way head in training. Balance is structural
rather than reweighted: the four-way head keeps only items where all three rules have a
candidate, and the binary head draws one illegal row per item. Fitting is multinomial logistic
regression with an $\ell_2$ penalty, optimized full-batch by L-BFGS (300 iterations,
strong-Wolfe line search) on features standardized by the training split's mean and standard
deviation. The penalty is chosen from $\{10^{-4}, \ldots, 1\}$ on validation macro-F1, selecting
$10^{-4}$ for both heads in the raw space. On the held-out test split the binary head reaches accuracy 0.8896 and
AUROC 0.9418, and the four-way head accuracy 0.7760 and macro-F1 0.7734, with R1 its weakest
class (F1 0.6402).

\textbf{Probe directions.} The five pooled directions are read off the two committed legality
probes: from the four-way head, $\mathrm{unit}(w_i - \bar{w})$ for legal, R1, R2 and R3; from the
binary head, $\mathrm{unit}(w_{\mathrm{illegal}} - w_{\mathrm{legal}})$. The per-square directions are the
same read-outs refit per candidate move square by ridge regression ($\lambda = 10$) on centered
activations and centered one-hot targets, one point per playable square, so each class
contributes $n = 60$ points. MDS takes in the matrix of geodesic angles $\theta_{ij} = \arccos(x_i^{\top} x_j)$ between unit-normalized probe directions, and places these geodesic angles in 2D space such that their Euclidean distances correlate with how close the directions are. We embed per-square points and pooled directions in one
two-dimensional non-metric MDS fit~\citep{kruskal1964mds} (SMACOF, deterministic classical-scaling initialization, one
restart, 1000 iterations) on this matrix of geodesic angles; we use the angle rather than $1 - \cos$ because it is a
metric, and the orientation is pinned by orthogonal Procrustes onto the PCA coordinates so a
refit cannot rotate or mirror the panel. The embedding preserves family membership but not finer
structure -- its rank fidelity, 0.504, sits below a structure-free null of $0.574 \pm 0.007$ --
so read it for which families separate, and read angles off the cosine-similarity matrix among
the five pooled directions. In $\mathbb{R}^{512}$, random unit directions meet at $|\cos| \approx
0.035$. Read those entries against that scale. In the MDS panel
the ring is the pooled committed probe, a separate fit on pooled rows, not the centroid of its
own points.

\textbf{What is held out.} The per-square refits use all rows of the fit set, so the MDS panel
describes the geometry of the fit rather than a held-out one. The board probes are fit on 20,000
games of the same seed-42 stream, so they have seen most of the games the bracket-flip events
come from; the accuracies quoted above are the probes' own held-out test numbers, and those
panels are statements about geometry rather than about generalization. The legality heads are
scored on the spliced branch tree below, which is disjoint from the 800 probe-fit games by exact
token match.

\subsection{Counterfactual constructions}

Three constructions manufacture a contrast: free-running rollouts, bracket-flip events with a
matched legal control, and a spliced branch tree that changes one token.

\textbf{Free-running rollouts.} We sample 1000 games from the model at temperature 1.0 --
ancestral, no top-$k$ or nucleus, with the pass token masked -- taking the first move uniformly
over the four legal openings and generating all 58 remaining tokens, at seed 42. Illegal moves
are kept rather than resampled; 885 of the 1000 rollouts are legal throughout. Each rollout is
then teacher-forced on its own tokens, and at every $t \in [0, 56)$ its next-move mass is split
by the taxonomy, keeping only positions where the rollout is still legal through $t$ and the
mover does not forfeit, which leaves 54,766 positions. The reported curve is the per-game
cumulative $1 - \prod_{t' \le t}(1 - p(t'))$ of that mass, averaged over the games still alive at
$t$, on a logarithmic $y$ axis. Because a game leaves the population at its first illegal move,
$n$ falls from 1000 to 891 across the axis, which is also why the curves are not monotone. This
is the probability composed from the model's own next-move distribution, not the rate at which
rollouts actually sample an illegal move, which is 11.5\% [9.5, 13.6] by ply 59.

\textbf{Bracket-flip events.} An \emph{event} is a move that flips exactly one disc: for prefix
$P = \texttt{tokens[:}t \texttt{]}$ with mover $e$, the played move $m$ has flip set $\{s\}$, so
the board before and after differ in exactly two squares. Positions where either side forfeits at
$t$ or $t+1$ are dropped, since a forfeit at $t+1$ leaves $e$ to move again and inverts the
prediction the study rests on. The matched control $m'$ is another legal move for the same mover
from the same position that leaves $s$ unflipped and does not forfeit the opponent, so it is
matched on prefix, move index, mover and next mover, and differs only in leaving $s$ alone. A
control is \emph{strict} if its placement and every disc it flips stay off all eight of $s$'s
rays, so the control branch cannot reach $s$ along any line.

The population read-out is 1000 strict-control events, 40 at each of the 25 move indices $t \in
[10, 34]$, sampled at seed 42. We use a fixed quota per move index rather than whatever the
corpus holds, because move index is the strongest nuisance variable here and it is the window the
legality and detector studies also rest on. Meeting that quota needs a larger mining run -- 4000
events, of which 3194 have a strict control -- from the same miner, stream and seed. That run scans a longer prefix of
the same stream than the 2000-event corpus every other measurement uses, and mines it
independently, so the event drawn for the single-game panel is not among these 1000. Each event has exactly one
flipped and one placed square but about 12 unflipped mover's and 13 unflipped opponent's squares,
so the five role markers average over 1000, 1000, 11,935, 13,065 and 37,000 square-observations.
Open markers are the state before $m$, filled markers the state after, and the cross is where
$m'$ leaves $s$; the dashed arrow to that cross is the panel's one contrastive element, and the
rest is definitional. Both axes are read in each condition's own mover-relative frame, and the
frame alternates with the ply, so every unflipped disc crossing $x = 0$ is a consequence of that
convention rather than a finding about the model. The claim the panel supports is the gap between
the two heads at $s$: $13.98$ logits [13.86, 14.10].

\textbf{The spliced branch tree.} Each source game is forked once at $t_0$ into up to nine branches
that share \texttt{tokens[:$t_0$]} and \texttt{tokens[$t_0$+1:]} and differ only at $t_0$: the
move actually played, two further legal moves, and one move per available illegal category, drawn
uniformly within its category at seed 42. There are two legal controls rather than one because a
second is what supplies the legal-versus-legal null. Nothing is regenerated after $t_0$ -- the
suffix is kept verbatim -- so a difference downstream comes from that one token and nothing
else. Fork points are round-robin over $t_0 \in [10, 34]$, and a later pass appended
games forked at $t_0 = 20$. The tree holds 3923 games and 34,546 branches, 8.81 per game: a game contributes a branch
only for the illegal categories its board admits. After $t_0$ a branch
replays the source game's suffix onto a board that no longer exists, so a trajectory's right half
shows the score decaying and is not a legality judgment of those moves.

\subsection{The detection statistic}

Equations (1)--(3) are computed exactly as written: the step length $d_t$, its standardization
$g_t$ against a per-move-index legal profile $\mu_t, \sigma_t$, and the windowed score $S_t$.
$g_t$ is the negative of the $z$-score, so that a \emph{shorter}-than-usual step scores high. With
the sink dropped, a game yields 57 steps, and the candidate move index is the step index plus two,
so an $m = 4$ window reaches moves 6--55: 50 candidates.

\textbf{Operating point.} The threshold $\tau = 0.618$ is the cut maximizing balanced accuracy,
found by an exact sweep of the breakpoints over all 2415 scored games rather than the 500 the
population panels draw, pooling every
scored position up to $t_0$ across all four branch blocks with positives only at $t_0$ of the
illegal arm. It gives TPR 0.658 and TNR 0.683 there. The objective is nearly flat -- balanced accuracy is
0.6644 at $\tau = 0.869$ -- so $\tau$ names a trade-off rather than a sharp optimum. Scored at
$t_0$ on the 500-game set, 37\% of legal controls still sit above it. The same procedure refit in the best-ranked dictionary space
gives $\tau = 1.245$.

\subsection{Sparse autoencoders} \label{app:SAE_descriptions}

We compare eight architectures at expansion factor 16, giving 8192 latents on
$d_{\mathrm{model}} = 512$, on layer-5 activations: Standard (ReLU)~\citep{bricken2023monosemanticity}, $p$-Anneal~\citep{karvonen2024measuring}, Gated~\citep{rajamanoharan2024gated},
Gated+$p$-Anneal~\citep{karvonen2024measuring}, TopK~\citep{makhzani2013ksparse, gao2025scaling}, BatchTopK~\citep{bussmann2024batchtopk}, the prototype-distance architecture SpaDE~\citep{hindupur2025projecting},
and CosineSpaDE, a variant of it we introduce here.

\textbf{Training.} Our own dictionaries see 300M tokens at batch 8192, i.e. 36,621 steps, at
learning rate $3\times10^{-4}$ ($2.83\times10^{-4}$ for the TopK family, which sets it from the
dictionary size), with 1000 warmup steps and seed 42. Activations are standardized before any autoencoder sees
them, $x' = \sigma (x - \mu)$ with $\mu$ the mean over the first 100 calibration batches and
$\sigma = \sqrt{d_{\mathrm{model}} / \mathbb{E}\|x - \mu\|^2}$, so $\mathbb{E}\|x'\|^2 = d_{\mathrm{model}}$. The same transform is applied at encode time. SpaDE scores prototypes by
$-\lambda\|x - Ae_i\|^2$ with $\lambda$ trainable through a softplus, encodes with sparsemax~\citep{martins2016sparsemax},
and adds the distance-weighted regularizer of K-Deep Simplex~\citep{tasissa2023kds},
$\gamma \sum_j \|x - Ad_j\|^2 f_j$; CosineSpaDE scores
by cosine similarity with no $\lambda$ and regularizes with $\sum_j (1 - \cos(x, Ad_j)) f_j$,
which is bounded in $[0, 2]$ because sparsemax sums to one, so its natural $\gamma$ scale is
orders of magnitude larger.

\textbf{Matching sparsity.} Trained at each family's own default hyperparameters, our eight land anywhere on the sparsity
axis, $L_0$ 32.5 to 1808, so a gap between two spaces confounds
architecture with sparsity. For the dictionary comparison, we instead hold every family at one
operating point. Each type first names its own best-scoring configuration on an equal-weight mean
of four min--max normalized metrics from Karvonen's board-game suite~\citep{karvonen2024measuring} (coverage,
reconstruction F$_1$, loss recovered, variance explained) over the 8192-latent half of a 516-checkpoint grid, seed-collapsed to 257
configurations; the mean of those
eight $L_0$ values, \textbf{219.16}, becomes a shared target, and each type is then re-seated at
whichever member of its own grid is nearest that target in $\log L_0$ -- by sparsity, not by
score. Realized $L_0$ runs from 183.79 to 260.13, a $1.42\times$ band. CosineSpaDE sits at the top of its own grid: a bounded cosine score with no inverse
temperature cannot leave $L_0 \in [16.1, 183.2]$ at any $\gamma$, which is what kept the target
from going higher.

\textbf{The roster.} Five of the eight seated checkpoints are Karvonen's released ones~\citep{karvonen2024measuring} (Standard \texttt{trainer25-v2}, $p$-Anneal \texttt{trainer22},
Gated \texttt{trainer32}, Gated+$p$-Anneal \texttt{trainer25}, TopK \texttt{trainer\_61}) and
three are ours (BatchTopK at $k = 220$, SpaDE at $\lambda$ initialized to $10^{-3}$ and
$\gamma = 5\times10^{-3}$,
CosineSpaDE at $\beta = 1, \gamma = 0.05$). Those five are trained on the raw residual stream,
whereas ours are trained on the standardized one, so each is encoded under the convention it was trained
in, and the two families' latents are not on one input scale: the roster compares architectures
at matched $L_0$, not one quantity measured eight ways. Matching is on sparsity only --
reconstruction F$_1$ across the eight still runs 0.464 to 0.938.

\textbf{Reading the detector in a dictionary.} The layer-5 activation at each position is encoded
to the dictionary's 8192 latents and the step length $d_t$ is taken there, on the latent
code and not on the reconstruction, with $\mu_t$ and $\sigma_t$ refit in that space on the
same held-out legal block. Same 500 games, same statistic, same profile construction.

\begin{figure}[htbp]
  \centering
  \includegraphics[width=0.8\linewidth]{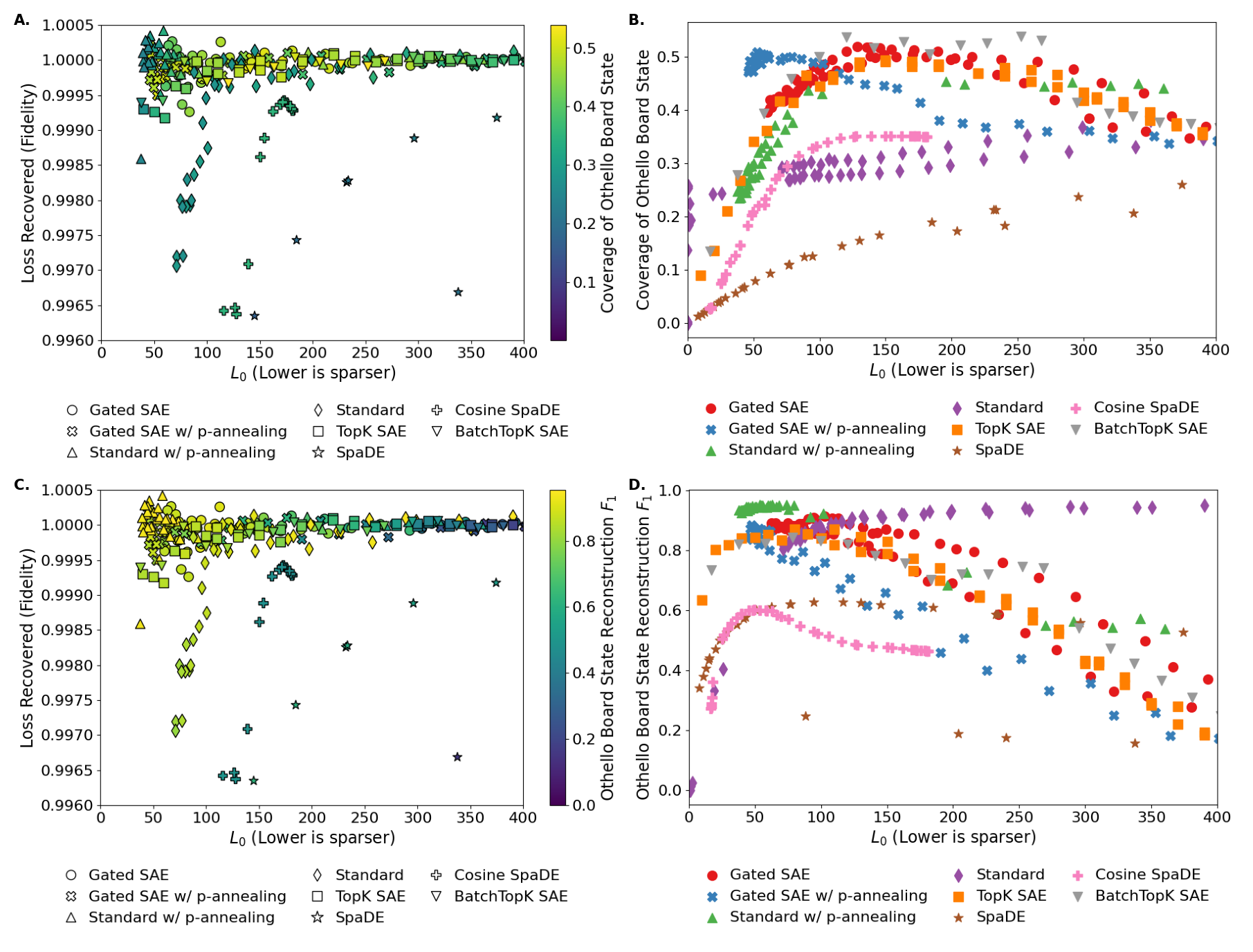}
  \caption{\textbf{Hyperparameter sweep of each SAE architecture used in this work across sparsity $L_0$.} Every point is one trained dictionary at expansion factor $16$ ($8192$ latents) on layer-$5$ activations, scored with Karvonen's board-game suite~\citep{karvonen2024measuring}; all four panels share the $x$-axis $L_0$, clipped to $[0, 400]$, and marker shape denotes the architecture. \textbf{(a)} Loss recovered (fidelity) against $L_0$, each point colored by its coverage of the Othello board state. \textbf{(b)} Coverage of the Othello board state against $L_0$. \textbf{(c)} As in (a), each point colored by its board state reconstruction $F_1$. \textbf{(d)} Board state reconstruction $F_1$ against $L_0$. }
  \label{fig:sae_sweep}
\end{figure}

\textbf{The evaluation suite.} Both the operating point above and the sweep in \cref{fig:sae_sweep} are
scored with Karvonen's board-game suite~\citep{karvonen2024measuring}, on 1000 held-out games. Let $f_i(x)$ be latent $i$'s
activation on a layer-5 activation $x$, let $f_i^{\max}$ be its maximum over a firing-rate pass,
and let $\phi_{i,\tau}(x) = \mathbf{1}[f_i(x) > \tau f_i^{\max}]$ binarize it at relative threshold
$\tau \in \{0, 0.1, \ldots, 1\}$. A \textit{board state property} $g_{r,c,k}$ is the binary event
that square $(r,c)$ holds class $k \in \{\rm mine, \rm blank, \rm theirs\}$, player-relative, at a
given position. Four metrics enter the score:

\begin{itemize}
\item \textbf{Sparsity} $L_0$, the mean number of nonzero latents per activation.

\item \textbf{Loss recovered (fidelity)}, $(H_{\rm SAE} - H_0)/(H_{\rm clean} - H_0)$, where
$H_{\rm clean}$ is the model's next-token cross-entropy, $H_{\rm SAE}$ is that cross-entropy with
the layer-5 residual stream replaced by the autoencoder's reconstruction $\hat{x}$, and $H_0$ is that cross-entropy
with that activation zeroed. Anchored at zero ablation this way, it is not confined to $[0, 1]$.

\item \textbf{Variance explained}, $1 - \sum_d \mathrm{Var}[(x - \hat{x})_d] / \sum_d \mathrm{Var}[x_d]$,
summed over the $d_{\mathrm{model}}$ coordinates.

\item \textbf{Coverage}, $\max_{\tau} \frac{1}{192} \sum_{r,c,k} \max_i F_1(\phi_{i,\tau}; g_{r,c,k})$:
at one threshold, the best single latent's $F_1$ for each property, averaged over properties. It
asks whether the dictionary \textit{contains} a latent that tracks a property, one latent at a time.

\item \textbf{Board state reconstruction} $F_1$. Latents are first labeled with the properties they
predict at precision $\ge 0.95$ on at least 10 positions; the board is then rebuilt at each position
as the union of the labels of every latent that fires there, and $F_1$ is micro-averaged over all
(position, square, class) cells. It asks whether the labeled dictionary \textit{jointly}
reconstructs the board.
\end{itemize}

The last two are easily conflated but are different tasks, not two aggregations of one number:
coverage macro-averages over properties and lets each pick its own best latent, while
reconstruction pools a single micro-averaged $F_1$ over cells and uses the whole labeled
dictionary at once. Masking the blank channel leaves 128 live properties of the 192 the suite
normalizes by, so Othello coverage cannot exceed $128/192 = 0.667$; this is why the coverage axis
of \cref{fig:sae_sweep} tops out near $0.5$ and not near $1$, and we report the uncorrected values so that
they stay comparable to~\citep{karvonen2024measuring}.


\subsection{Evaluation protocol}

\textbf{Scored populations.} We score every branch-complete game with $t_0 \in [10, 34]$ that
either is absent from the legal corpus below or sits at an odd index in it, which is what keeps a
scored trajectory out of the profile it is scored against: 2415 games, 21,735 branch rows. The
per-move-index profile $\mu_t, \sigma_t$ is fit on a block of 2000 legal games, minus the 778
that are themselves scored here -- a trajectory must not contribute to the baseline it is scored
against -- leaving \textbf{1222}. The evaluation set for the population panels is 500 games, an
equal quota of 20 at each of the 25 values of $t_0$, taken as the first by sorted game id. The
tree is not balanced on move index, forking 1665 of the eligible games at $t_0 = 20$, and move
index is the strongest nuisance variable it has, which is what the quota removes. Those 500 games
are written down once and re-read by the legality-probe AUROC bars, the residual-stream detector
panels and the dictionary ranking rather than re-dealt, so those three score one set of games
rather than three built by the same recipe.

\textbf{Scoring.} Every AUROC is taken at each row's own $t_0$ -- one value per branch, not a
maximum over the window -- for the illegal branch against the same source game's legal control at
the same $t_0$. A rule arm is one branch per game drawn at seed 42, never a mean over its member
categories, so R2 with three members carries the same estimator as R3 with one. AUROC is
rank-based with ties averaged. For the probe heads the score is $1 - P(\text{legal})$, taken from
the binary head for the first bar and from the four-way posterior for the rule bars; for the
detector it is $S_t$. The dictionary ranking is nine bars -- raw layer 5 and the eight
$L_0$-matched dictionaries -- sorted by height and ranked on the same at-$t_0$ paired statistic,
which is why raw reads 0.681 in both the residual and the dictionary panel.

\textbf{Chance.} Chance is measured rather than assumed: a game's legal control scored against
the second legal control the same game carries. For the probe heads, this gives 0.4875 [0.4587,
0.5144], which licenses the 0.5 line; across dictionary spaces it lands in $[0.485, 0.508]$.

\textbf{Density panels.} Class distributions are Gaussian kernel density estimates (Scott's rule)
rescaled to a count axis on one slice width shared across all five series and across spaces, so a
curve's height is comparable between panels. The dictionary density panel is structurally forced
to draw the winner of the ranking above.

\textbf{Uncertainty.} Every interval is a 95\% percentile bootstrap, computed and written out
even where the panel does not draw it -- the rollout curve carries five series and reads better
without bands. The resampling unit
is the source game, except for the bracket-flip population, where it is the event, and for the
probe-head AUROC bars, where it is the game pair. The rollout curve uses 2000 resamples at seed
0; the bracket-flip bars use 2000 resamples at seed 42; every AUROC bar uses 1000 resamples at
seed 42. The bracket-flip bars are a cluster bootstrap on the \emph{paired} displacement within
event, which is what an arrow asserts -- the endpoints move together, so an endpoint interval
understates the arrow by 1.4 to 1.8$\times$ on $x$.

\subsection{Illustrative positions}

Three single-game board panels are chosen for legibility, not sampled. The selections are
mechanical, but the criteria are about what reads in a picture, so these are illustrative games
rather than typical ones; the populations carry the claims.

\textbf{The rule panel.} Game 606 of the seed-42 stream at $t = 16$, black to move. We scan $t
\in [14, 26]$ over the first 800 games for a position where all three rules have a candidate, the
played move flips discs in more than one direction, and the four drawn squares stay at least
three cells apart. All three rules co-occur at 97.8\% of positions, so nothing here is rare. The
played move b4 is the legal one, and the violations drawn are \texttt{occupied\_opp} at b2 (R1),
\texttt{empty\_no\_bracket} at h3 (R2) and \texttt{opponent\_only} at f7 (R3). R2 is drawn from
its deepest member because the shallowest, \texttt{empty\_detached}, is an empty square with
nothing to see.

\textbf{The bracket-flip panel.} Game 550 at $t = 25$, white to move, with $s = $ c6, $m = $ d7
and $m' = $ g1. Among strict-control events at the median move index, we take the one whose
flipped-square displacement is closest to the median displacement over all 2000 events (pool of
34; median 9.566, this event 9.232). It is typical in the measurement, not in the picture, and
$t = 25$ is the corpus median rather than a chosen value.

\textbf{The detector trajectory.} Game 3889 at $t_0 = 20$. It is chosen so that all three rules
have a branch there, $t_0 \pm 5$ stays inside the scanned band, the score peaks at $t_0$ on all
three rule arms while the matched legal control does not, and the detector is right about at
least four of the five plies drawn; among those, the largest margin over its own runner-up. The
last two are selections over games. Off $t_0$ the row scored is the reference game, whose move
there is the legal one it played, so anything above $\tau$ is a false alarm; at $t_0$ it is the
drawn branch, and the rule members drawn are \texttt{occupied\_opp},
\texttt{empty\_no\_bracket} and \texttt{opponent\_only}. The shaded region is $t \le t_0 - m$,
where all four branches are still bit-identical because the window has not yet reached the
splice.

\section{Further Results}

In this section, we compare the rule violation metric we selected with other candidate metrics. We also present additional results
that we did not include in the main paper.


\begin{figure}[htbp]
  \centering
  \includegraphics[width=0.8\linewidth]{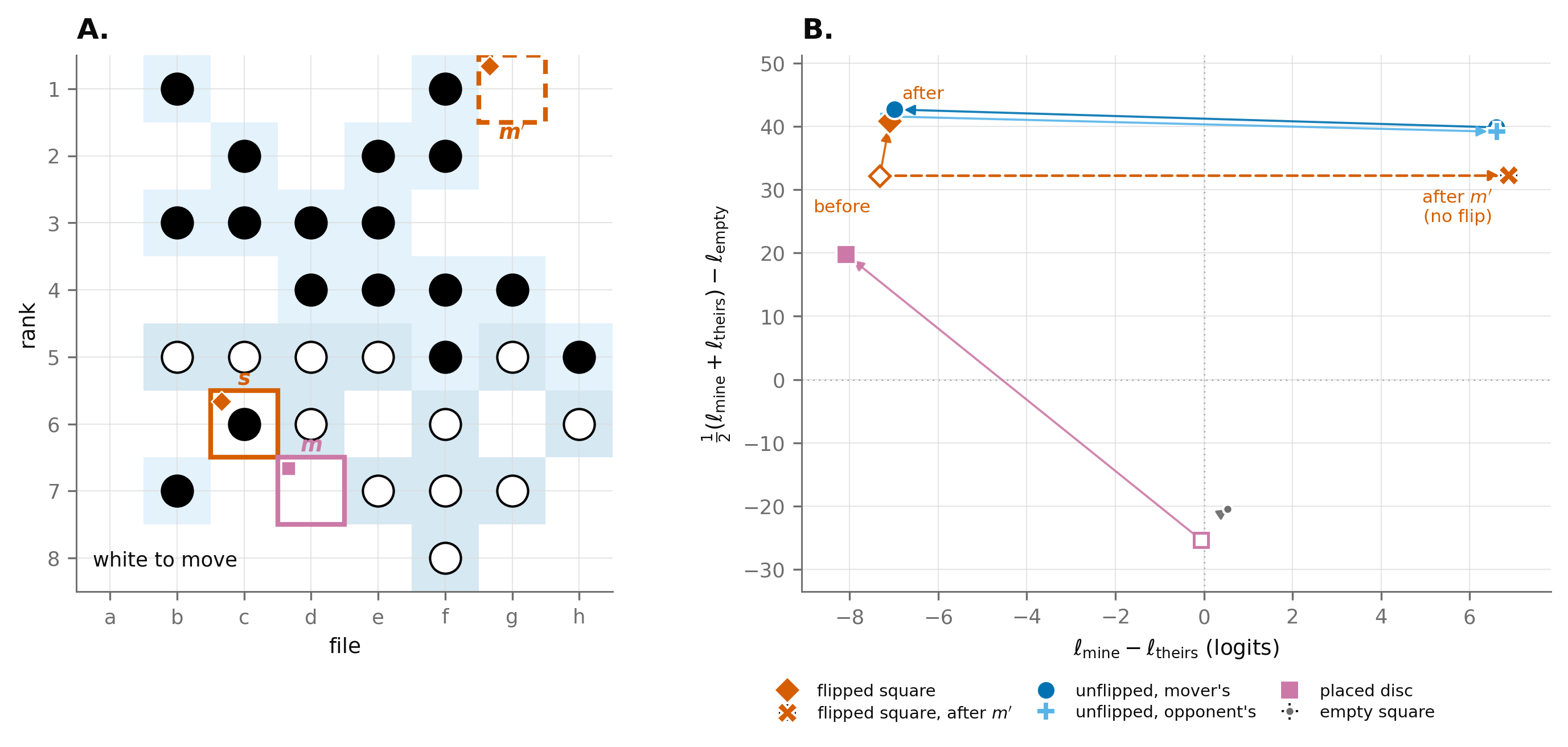}
  \caption{\textbf{Concrete example of OthelloGPT following the rules.} \textbf{(a)} Randomly chosen game, board state at move $t = 25$, white to move. `m' denotes move $t = 26$, `s' denotes the opponent's flanked disc, and `$m'$' denotes the matched control move. \textbf{(b)} Tensor product probe (TPR) on the board state of Othello. The y-axis indicates whether or not a square is occupied, and the x-axis indicates OthelloGPT's understanding of the relative disc color ($\rm mine$ versus $\rm their$ directions). The TPR shows that OthelloGPT updates its internal board representation in accordance with the flanking rule, changing the represented color of the flanked disc after the move.}
  \label{fig:flanking}
\end{figure}


\subsection{Rule Violation Metric Comparison Experiments}

The detector reads one scalar per position and asks whether it departs from what a legal game
does at the same move index. Step length is not the only candidate for that scalar, so we
compare it against the two alternatives a reader is most likely to prefer. Write $z_p$ for the
representation at position $p$ -- the layer-5 residual stream, or a dictionary's latent code --
and $\Delta_p = z_{p+1} - z_p$ for the displacement between consecutive positions.

\begin{itemize}
\item \textbf{Step length,} $d_p = \|\Delta_p\|_2$, the statistic this work uses. It is a property
of the \textit{pair} $(z_p, z_{p+1})$ and carries the magnitude of the move between them.

\item \textbf{Representation norm,} $\|z_p\|_2$, the norm of the representation itself, which in a
dictionary is the norm of the latent code and not of the reconstruction. It is a property of a
\textit{single} position, so alone among the three it carries no displacement at all.

\item \textbf{Curvature angle,} $\theta_p = \arccos\!\left(\frac{\langle \Delta_p, \Delta_{p+1}
\rangle}{\|\Delta_p\|_2\,\|\Delta_{p+1}\|_2}\right)$, in degrees, the turn between consecutive
displacements. It divides both lengths out, so it is scale-free where $d$ is not, and it is a
property of the \textit{triple} $(z_p, z_{p+1}, z_{p+2})$: it needs two consecutive
displacements where $d$ needs one, and is undefined wherever either has zero length.
\end{itemize}

\begin{figure}[htbp]
  \centering
  \includegraphics[width=0.8\linewidth]{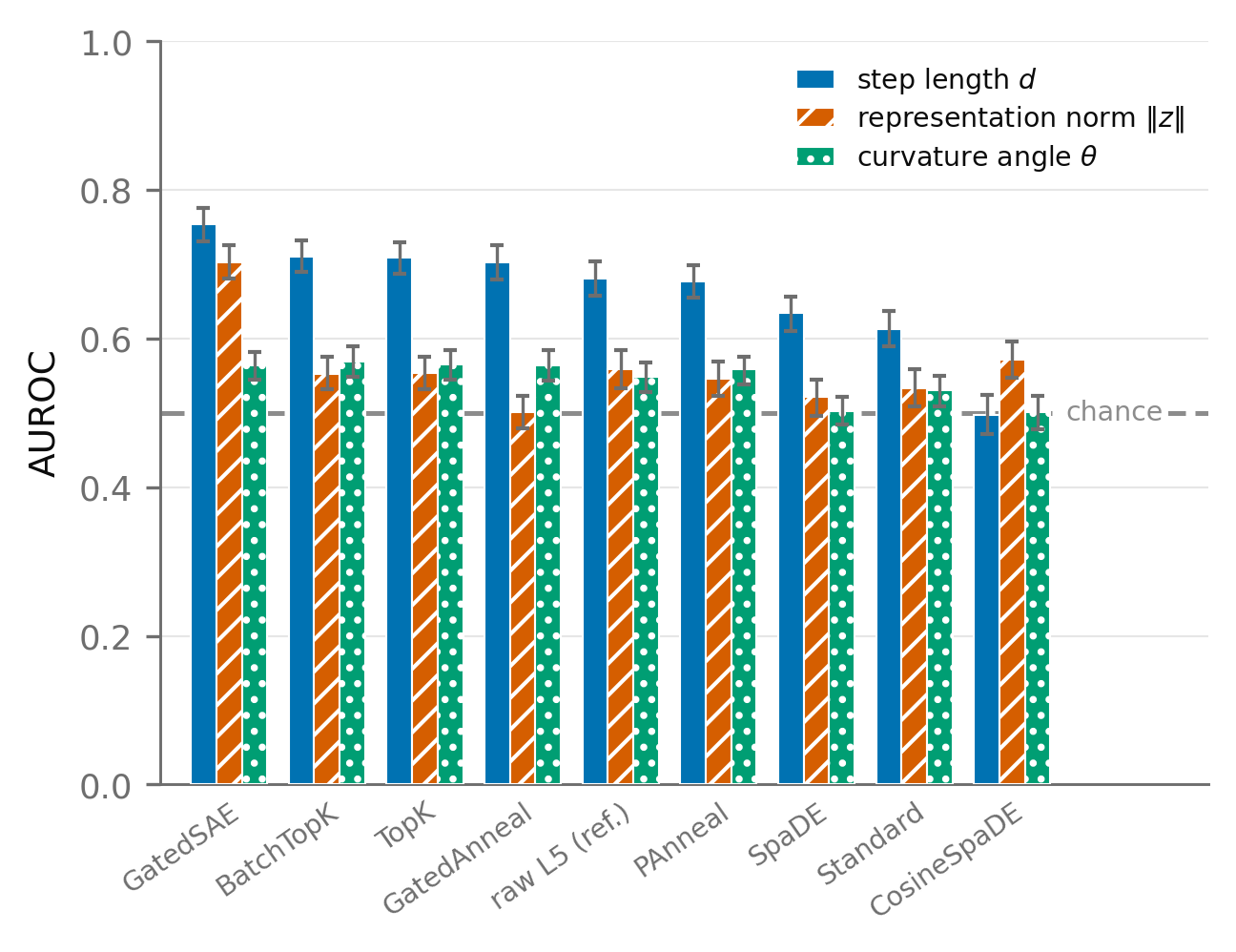}
  \caption{\textbf{Step length performs the best compared to the other rule violation detection metrics we tested.} AUROC comparing three rule violation detection metrics -- step length $d$, representation
  norm $\|z\|$, and turning angle $\theta$ -- in \textit{residual stream activations} and in
  the \textit{SAE latent activations}, averaged over
  $n = 500$ games.}
  \label{fig:detector_comparison}
\end{figure}

The three therefore differ in what they can see. The norm cannot see motion, the angle cannot
see how far the representation traveled, and only $d$ sees both a displacement and its size.
Each is standardized per move index against a $\mu_t, \sigma_t$ profile fit on the held-out
legal block, so one cut is meaningful across games, and each is read on the same $n = 500$
games at each game's own $t_0$. \Cref{fig:detector_comparison} reports the result.

Step length wins in eight of the nine spaces, and the turning angle exceeds $0.569$ nowhere.
The comparison is not tilted in its favor: $d$ is held at the direction this work fixed for it
-- shorter than the legal profile counts as illegal -- while the norm and the angle are each
granted whichever sign scores better, which is the one asymmetry here, and it favors the
challengers. Neither survives it. The norm's better direction is positive in six of the nine
spaces and negative in three, and a label-free detector has to fix its direction in advance, so
a quantity whose sign moves from space to space is unusable however well it scores.


\subsection{Rule Violation Diverging Trajectories Experiments}

In this section, we present two additional results. First, we show how activations in the residual stream and in the SAE latent spaces diverge after a rule-violating move. Second, we show how that violation changes their distance to the reference game. Both vary the same two things.

\begin{figure}[htbp]
  \begin{minipage}[t]{0.48\linewidth}
    \vspace{0pt}
    \centering
    \includegraphics[width=\linewidth]{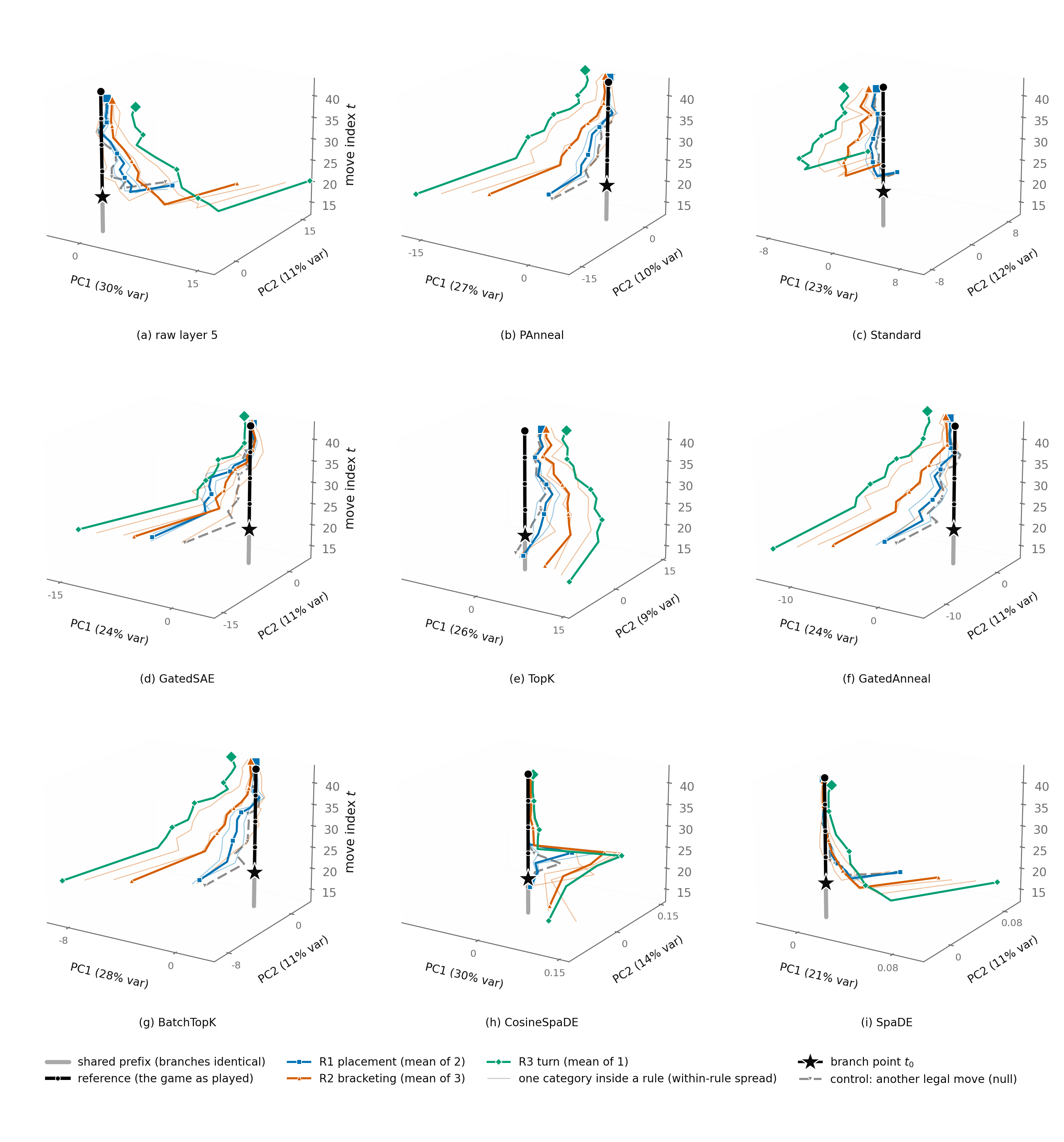}
    \caption{\textbf{Divergence of activations in the residual stream after a rule-violating move
    (moves are NOT regenerated after $t_0$).} Trajectories of one game's eight branches after
    $t_0 = 20$, projected onto the first two PCs of the displacement from the reference,
    $\Delta(t) = z_b(t) - z_{\rm ref}(t)$, with move index $t$ on the vertical axis.
    \textbf{(a)} \textit{Residual stream activations} at layer $5$, \textbf{(b)}--\textbf{(i)}
    \textit{SAE latent activations} of the eight SAE types. Branches share both the prefix and
    the suffix with the reference, so two of them differ in exactly one token.}
    \label{fig:traj_no_regen}
  \end{minipage}\hfill
  \begin{minipage}[t]{0.48\linewidth}
    \vspace{0pt}
    \centering
    \includegraphics[width=\linewidth]{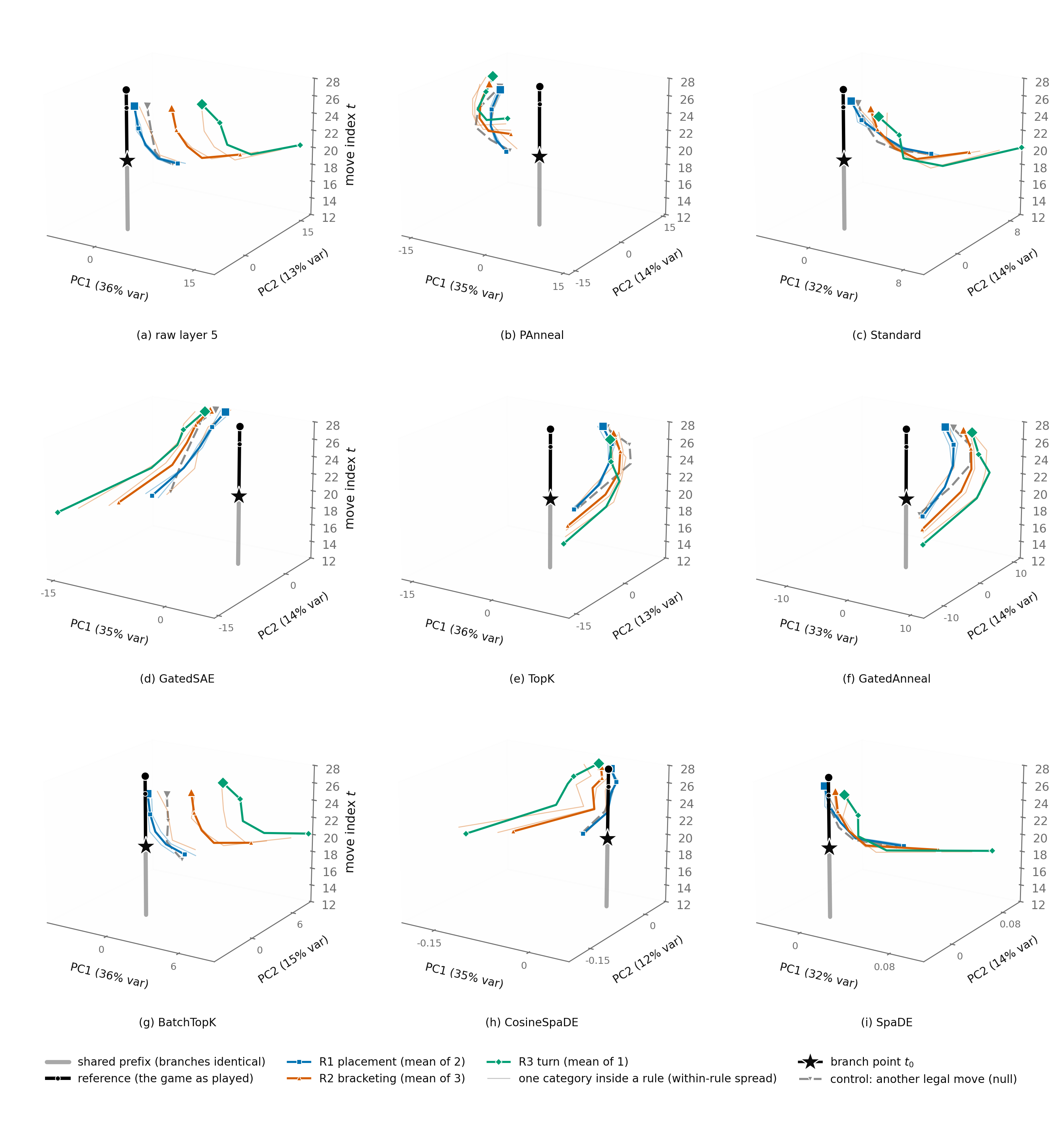}
    \caption{\textbf{Divergence of activations in the residual stream after a rule-violating move
    (moves ARE regenerated after $t_0$).} As \cref{fig:traj_no_regen}, but each branch samples its own
    continuation from OthelloGPT assuming the move at $t_0$ was played. Each path is the mean
    over $K = 4$ sampled futures, and the window stops at $t_0 + 8$, where the comparison
    saturates.}
    \label{fig:move_generation0}
  \end{minipage}
  
  \vspace{\floatsep}
  
  \begin{minipage}[t]{0.48\linewidth}
    \vspace{0pt}
    \centering
    \includegraphics[width=\linewidth]{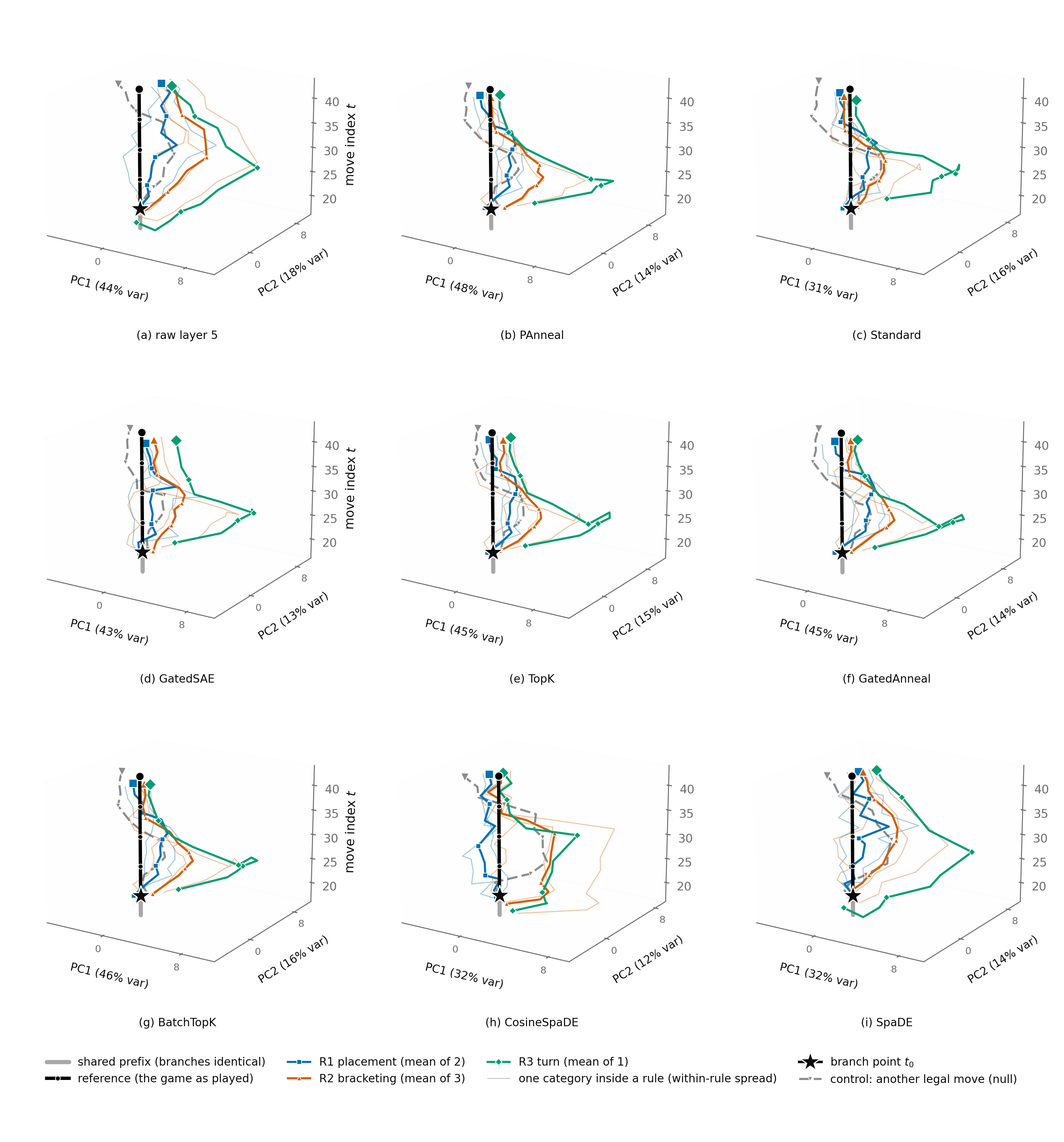}
    \caption{\textbf{Divergence of $R^{21}$ curvature descriptors after a rule-violating move
    (moves are NOT regenerated after $t_0$).} As \cref{fig:traj_no_regen}, but the PCs are of the curvature
    displacement $\Delta\varphi(t) = \varphi(z_b(t)) - \varphi(z_{\rm ref}(t))$, inside
    ReStraV's $21$-dimensional descriptor space of $11$ step-length and $10$ turning-angle
    components.}
    \label{fig:move_generation1}
  \end{minipage}\hfill
  \begin{minipage}[t]{0.48\linewidth}
    \vspace{0pt}
    \centering
    \includegraphics[width=\linewidth]{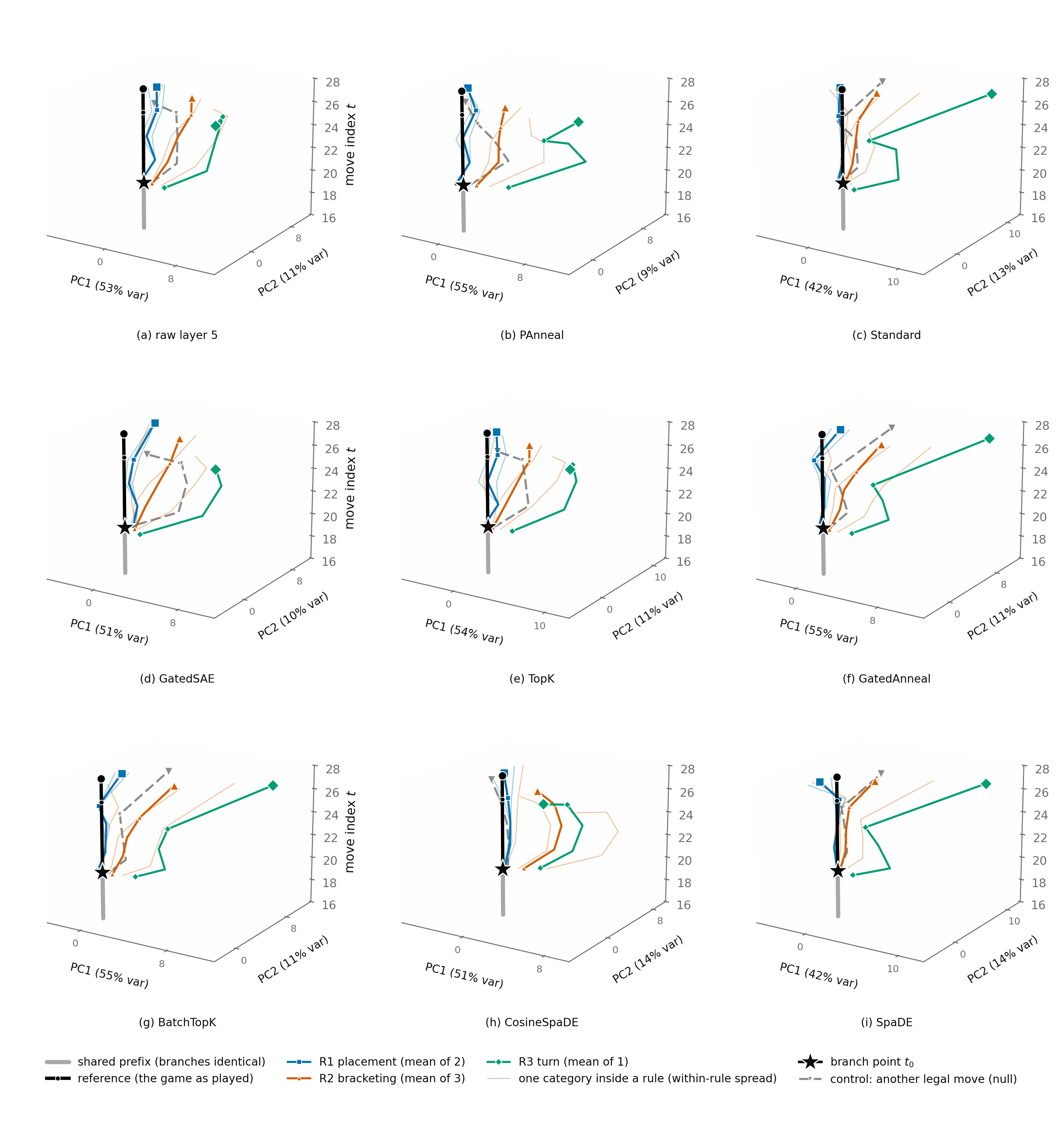}
    \caption{\textbf{Divergence of $R^{21}$ curvature descriptors after a rule-violating move
    (moves ARE regenerated after $t_0$).} \Cref{fig:move_generation1}'s descriptor plane on \cref{fig:move_generation0}'s
    regenerated branches.}
    \label{fig:move_generation2}
  \end{minipage}
  \end{figure}

\textbf{Whether the game is regenerated after $t_0$.} In the \textit{spliced} branch set, every
branch replays the source game's own remaining moves, so two branches are token-identical
everywhere except at $t_0$, and any later difference is attributable to that one token. The price
is that the shared suffix is a legal continuation of the reference's board alone, which no
illegal branch's board admits, so every branch but the reference carries a future it could not
have played. The \textit{rollout} set removes that confound: from $t_0$ onward every branch --
the reference included -- samples its own continuation from OthelloGPT at temperature $1.0$,
$K = 4$ samples per branch, with rollout $k$ of every branch drawn against the same uniform
stream so that the $K$ are paired and each path is the mean of the $K$ paired differences from
the reference. Only the pass token is masked, so a sampled continuation may itself be illegal;
that is the model's own behavior and is the object of study. What it costs is attribution, since
two branches now differ at $t_0$ and at every position after it, so neither set replaces the
other. \Cref{fig:traj_no_regen} and \cref{fig:move_generation1} are spliced, and \cref{fig:move_generation0} and \cref{fig:move_generation2} are their
regenerated twins, drawn only to $t_0 + 8$ because past that offset the illegal and control
displacements agree to within $0.5\%$ and cross over by $t_0 + 13$.

\textbf{Which space the trajectory is read in.} \Cref{fig:traj_no_regen} and \cref{fig:move_generation0} embed the
displacement $\Delta(t) = z_b(t) - z_{\rm ref}(t)$ in the activations themselves. \Cref{fig:move_generation1}
and \cref{fig:move_generation2} instead map each trajectory into ReStraV's $21$-dimensional curvature
descriptor~\citep{interno2025restrav} and difference it there. Over a trailing window of eight sampled positions, write
$\Delta z_i = z_{i+1} - z_i$, $d_i = \|\Delta z_i\|_2$, and
$\theta_i = \arccos\!\left(\frac{\langle \Delta z_i, \Delta z_{i+1}\rangle}{\|\Delta z_i\|_2\,
\|\Delta z_{i+1}\|_2}\right)$ in degrees; the descriptor is
\[
\varphi(z) \;=\; \big[\, d_1, \ldots, d_7,\; \theta_1, \ldots, \theta_6,\; m(d),\; m(\theta) \,\big]
\;\in\; \mathbb{R}^{21},
\]
where $m(x) = (\mathrm{mean}, \min, \max, \mathrm{var})$ over the whole sequence. It separates how
far a trajectory travels from how sharply it turns, which the raw displacement mixes together.
$\varphi$ is nonlinear, so it is applied to each branch on its own, and the difference
$\Delta\varphi(t) = \varphi(z_b(t)) - \varphi(z_{\rm ref}(t))$ is taken inside the descriptor: a
mean taken before $\varphi$ would measure the curvature of a path no branch travels. The trailing
window costs the first offsets, so these panels begin later than their activation counterparts.

\begin{figure}[htbp]
  \begin{minipage}[t]{0.48\linewidth}
    \vspace{0pt}
    \centering
    \includegraphics[width=\linewidth]{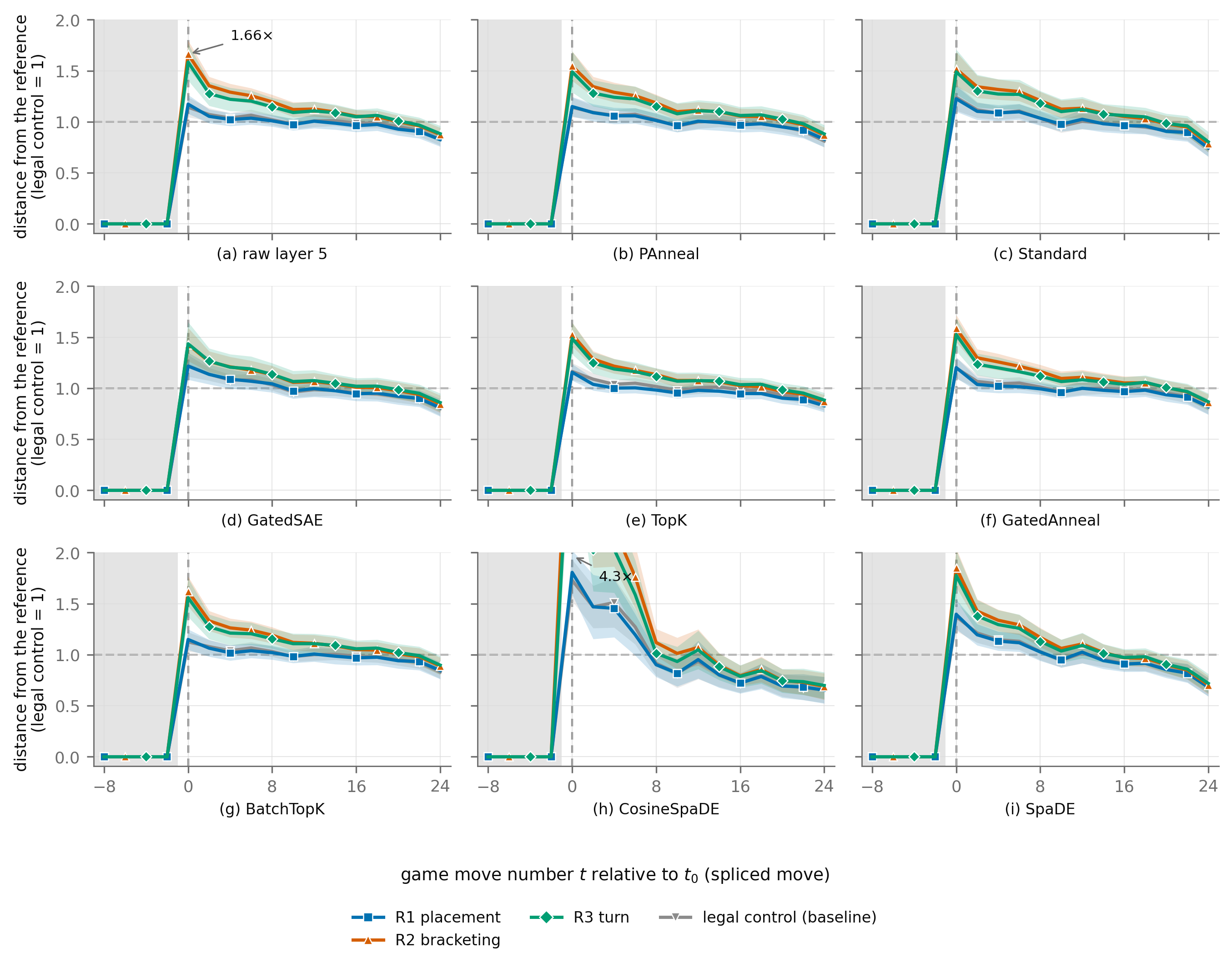}
    \caption{\textbf{Distance from the reference after a rule-violating move (moves are NOT
    regenerated after $t_0$).} Each rule's distance from the reference branch, measured in the
    plane its panel of \cref{fig:traj_no_regen} draws and divided by that panel's own legal control, against
    move index $t$ relative to $t_0 = 20$. Panels are as in \cref{fig:traj_no_regen}. One curve per rule, the
    mean of its categories; bands are $95\%$ CIs over $n = 100$ games. Left of $t_0$, the branches
    share their prefix and the distance is $0$ by construction; a curve that leaves the $y$ cap
    carries the value it reaches.}
    \label{fig:move_generation3}
  \end{minipage}\hfill
  \begin{minipage}[t]{0.48\linewidth}
    \vspace{0pt}
    \centering
    \includegraphics[width=\linewidth]{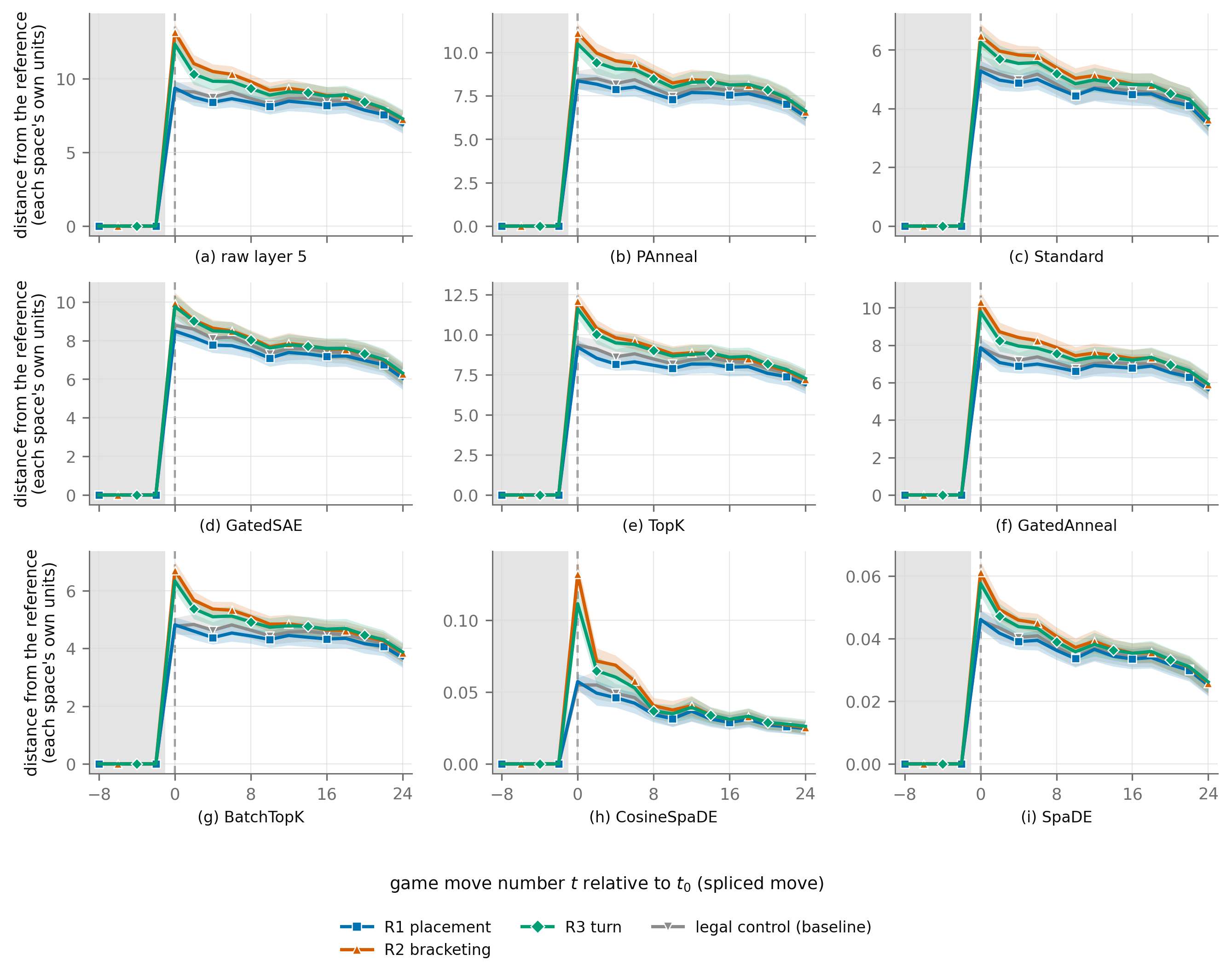}
    \caption{\textbf{Distance from the reference after a rule-violating move, unnormalized
    (moves are NOT regenerated after $t_0$).} \Cref{fig:move_generation3} without the divisor: distances in
    each space's own units, so every panel carries its own $y$ axis. What the nine draw spans
    $1360\times$, from SpaDE to raw layer $5$.}
    \label{fig:move_generation4}
  \end{minipage}
  
  \vspace{\floatsep}
  
  \begin{minipage}[t]{0.48\linewidth}
    \vspace{0pt}
    \centering
    \includegraphics[width=\linewidth]{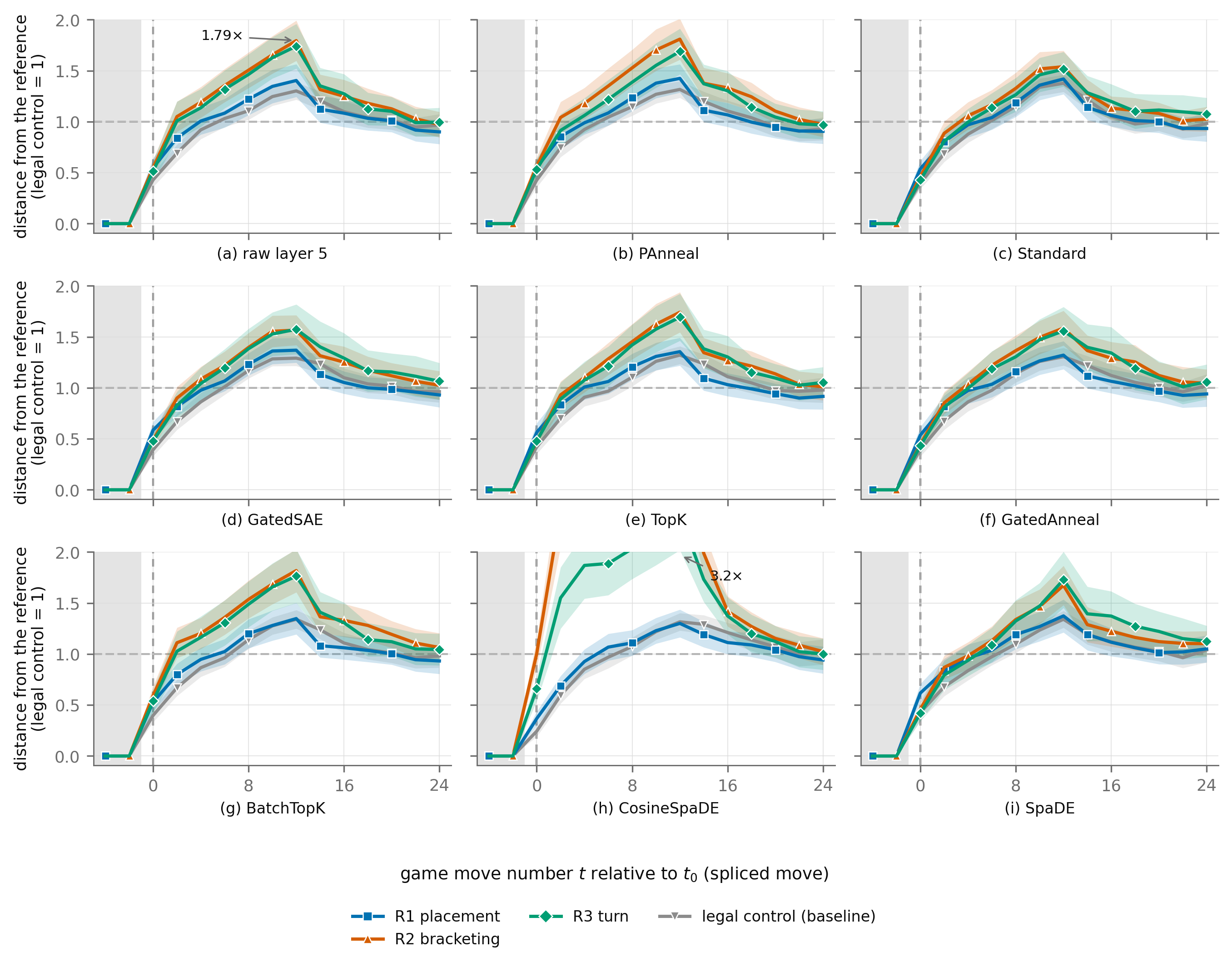}
    \caption{\textbf{Distance from the reference in $R^{21}$ curvature space after a rule
    violating move (moves are NOT regenerated after $t_0$).} As \cref{fig:move_generation3}, but distances are
    taken in \cref{fig:move_generation1}'s curvature descriptor plane rather than in the displacement plane.}
    \label{fig:move_generation5}
  \end{minipage}\hfill
  \begin{minipage}[t]{0.48\linewidth}
    \vspace{0pt}
    \centering
    \includegraphics[width=\linewidth]{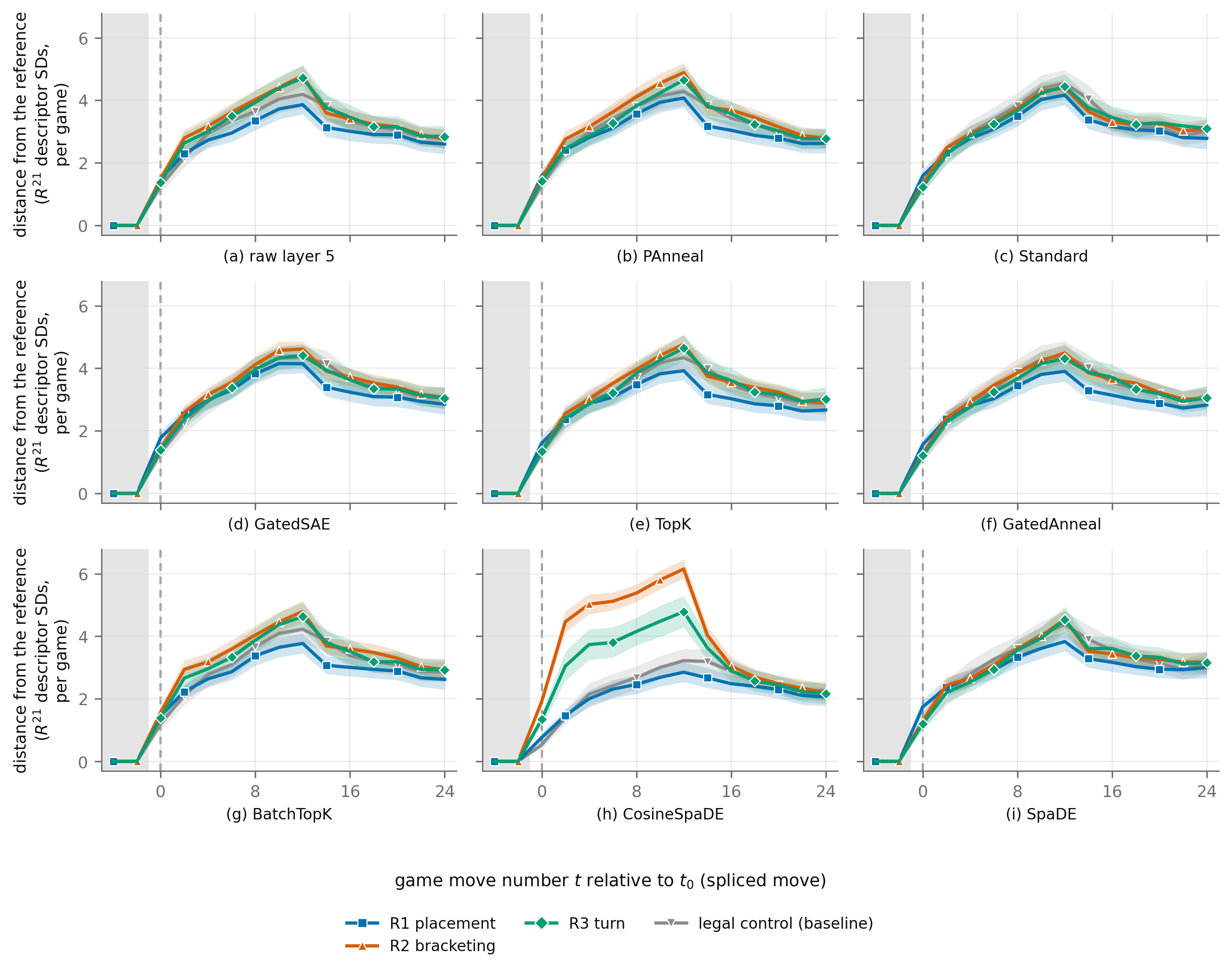}
    \caption{\textbf{Distance from the reference in $R^{21}$ curvature space, unnormalized
    (moves are NOT regenerated after $t_0$).} \Cref{fig:move_generation5} without the divisor, in per-game
    descriptor SDs. Column scaling puts every panel's peak within $1.3\times$ of every other's,
    so one $y$ axis holds all nine.}
    \label{fig:move_generation6}
  \end{minipage}
  \end{figure}

\textbf{Distance to the reference.} \Crefrange{fig:move_generation3}{fig:move_generation6} carry the magnitude a projection
cannot show -- how far each rule runs from the reference at every offset, in the same plane the
corresponding panel of \cref{fig:traj_no_regen} or \cref{fig:move_generation1} draws -- and we plot it in two forms.
\Cref{fig:move_generation3} and \cref{fig:move_generation5} divide by the legal control's own mean over $t \ge t_0$, one
scalar per panel and game, so the control keeps its shape and only averages to $1$. That is the
only form in which nine planes refit per space may be read side by side, since the control's own
in-plane distance differs by $\sim 700\times$ across the grid. \Cref{fig:move_generation4} and
\cref{fig:move_generation6} drop the divisor and draw those measured distances, which the ratio hides: in the
activations the nine spaces span $1360\times$, so each panel there carries its own $y$ axis,
while the descriptor's columns are scaled to unit variance per game and every panel's peak falls
within $1.3\times$ of every other's.

  
\end{document}